\documentclass[a4paper,fleqn]{cas-dc}

\usepackage[numbers]{natbib}

\def\tsc#1{\csdef{#1}{\textsc{\lowercase{#1}}\xspace}}
\tsc{WGM}
\tsc{QE}
\tsc{EP}
\tsc{PMS}
\tsc{BEC}
\tsc{DE}

\usepackage{amsmath,amsfonts}
\usepackage{algorithmic}
\usepackage{algorithm}
\usepackage{array}
\usepackage[caption=false,font=normalsize,labelfont=sf,textfont=sf]{subfig}
\usepackage{textcomp}
\usepackage{stfloats}
\usepackage{url}
\usepackage{verbatim}
\usepackage{graphicx}
\usepackage{cite}
\usepackage{soul}
\usepackage{booktabs}
\usepackage{pifont}
\usepackage{multicol}
\usepackage{multirow}
\usepackage{xcolor}
\usepackage{listings}
\usepackage{xspace}

\lstdefinelanguage{PDDL}{
    morekeywords={define, domain, requirements, types, predicates, action, parameters, precondition, effect, and, not},
    sensitive=true,
    morecomment=[l]{;},
    morestring=[b]"
}

\begin{document}
\let\WriteBookmarks\relax
\def\floatpagepagefraction{1}
\def\textpagefraction{.001}


\shortauthors{W. Akram et~al.}

\title [mode = title]{AquaChat: An LLM-Guided ROV Framework for Adaptive Inspection of Aquaculture Net Pens}

\author[1]{Waseem Akram}
    \address[1]{Khalifa University Center for Autonomous Robotic Systems (KUCARS), Khalifa University, United Arab Emirates.}
    \ead{waseem.akram@ku.ac.ae}
    \credit{Conceptualization, Methodology, Software, Writing - Original Draft, Formal Analysis, Writing - Reviews and Editing}

 \author[1]{Muhayy Ud Din}
     \credit{Methodology, Software - Original Draft, Investigation }
     \ead{muhayyuddin.ahmed@ku.ac.ae}
     
 \author[1]{Abdelhaleem Saad}
     \credit{Methodology, Software - Original Draft, Investigation }
     \ead{abdelhaleem.saad@ku.ac.ae}


\author[1]{Irfan Hussain}
     \credit{Methodology, Writing - Original Draft, Investigation }
     \cortext[cor1]{Corresponding author: I. Hussain (irfan.hussain@ku.ac.ae)}
\begin{abstract}
Inspection of aquaculture net pens is essential for maintaining the structural integrity, biosecurity, and operational efficiency of fish farming systems. Traditional inspection approaches rely on pre-programmed missions or manual control, offering limited adaptability to dynamic underwater conditions and user-specific demands. In this study, we propose AquaChat, a novel Remotely Operated Vehicle (ROV) framework that integrates Large Language Models (LLMs) for intelligent and adaptive net pen inspection. The system features a multi-layered architecture: (1) a high-level planning layer that interprets natural language user commands using an LLM to generate symbolic task plans; (2) a mid-level task manager that translates plans into ROV control sequences; and (3) a low-level motion control layer that executes navigation and inspection tasks with precision. Real-time feedback and event-triggered replanning enhance robustness in challenging aquaculture environments. The framework is validated through experiments in both simulated and controlled aquatic environments representative of aquaculture net pens. Results demonstrate improved task flexibility, inspection accuracy, and operational efficiency. AquaChat illustrates the potential of integrating language-based AI with marine robotics to enable intelligent, user-interactive inspection systems for sustainable aquaculture operations.

\end{abstract}


\maketitle

\section{Introduction}


Aquaculture has emerged as a critical component of global food security, meeting the rising demand for seafood while reducing pressure on wild fish stocks \citep{subasinghe2009global}. However, maintaining aquaculture net pens used for fish farming presents significant challenges. Structural integrity is essential to prevent fish escapes, minimize biofouling, and ensure the overall health of farmed species. Traditional net pen inspections often rely on manual divers or remotely operated vehicles (ROVs) controlled by human operators \citep{akram2021visual}. These methods can be labor-intensive, time-consuming, and prone to errors, especially in harsh underwater environments with poor visibility or complex structures. As the scale and complexity of aquaculture operations increase, there is a growing need for more intelligent, autonomous inspection solutions \citep{paspalakis2020automated}.

Marine robotics has emerged as a key enabler for aquaculture automation, providing tools and systems capable of performing complex underwater tasks \citep{sohan2024review}. 
ROVs are widely used in aquaculture for underwater inspection and maintenance. These vehicles can operate continuously in environments where human intervention is limited, offering significant advantages in efficiency and safety \citep{akram2023autonomous}. Traditionally, these vehicles rely on human operators to control their movements and interpret sensor data, which can be inefficient and subject to human error. Advancement in Artificial Intelligence (AI) and machine learning have further enhanced the capabilities of marine robots, allowing them to perform complex tasks such as navigation, object recognition, and defect detection with minimal human intervention. Enhancing ROVs with autonomous capabilities can improve their effectiveness and reduce operational costs. Modern ROV systems can incorporate advanced navigation and sensing technologies to perform tasks independently, with minimal human oversight \citep{salin2018aquaculture}.

\begin{figure}[t]
    \centering
    \includegraphics[width=1\linewidth]{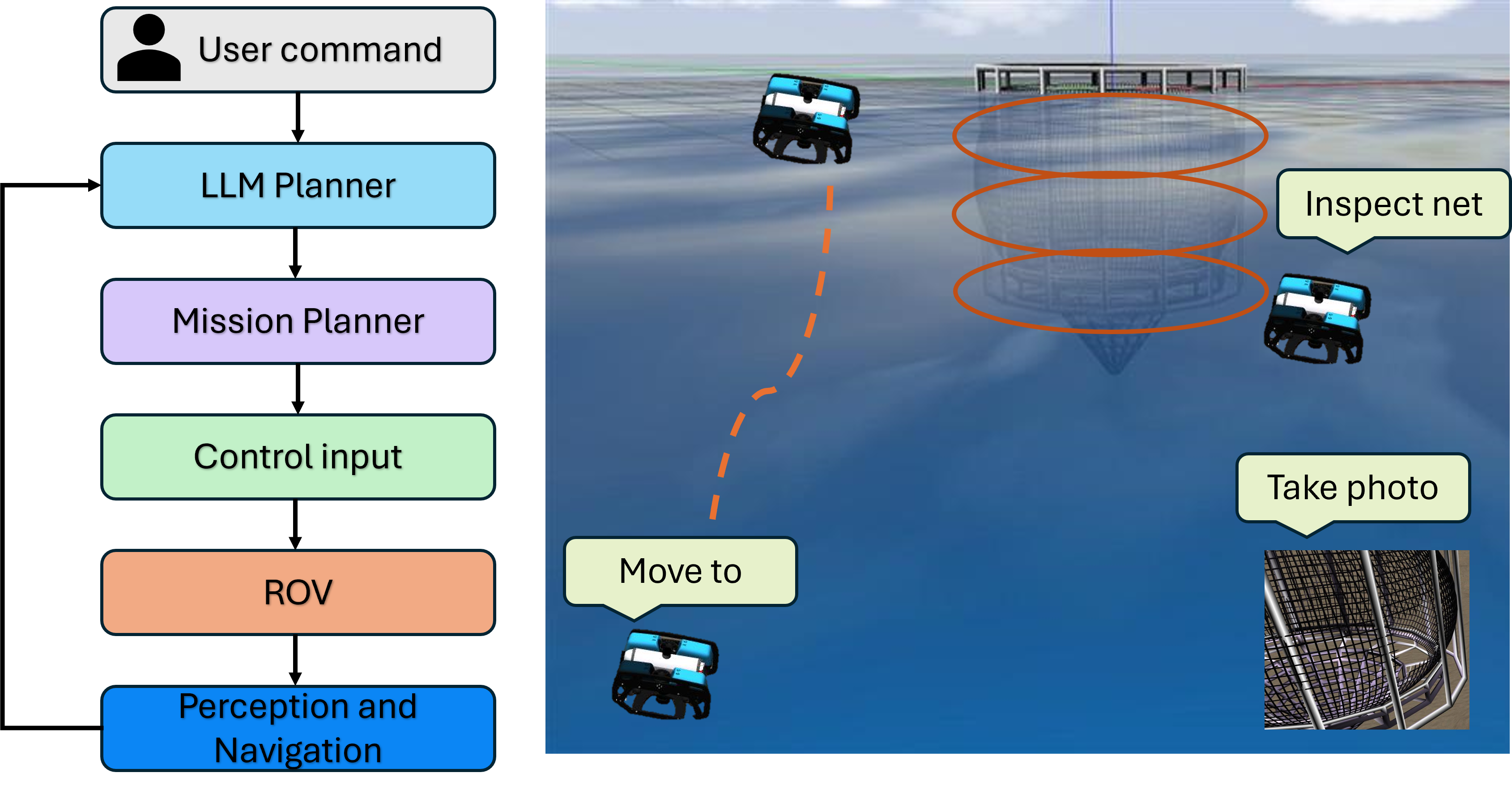}
    \caption{AquaChat Project Overview: Using LLM to interpret user commands and guide adaptive ROV navigation and inspection of aquaculture net pens. The system integrates a low-level motion planner to execute action sequences such as navigation, inspection, and data capture.}
    \label{fig:workflow}
\end{figure}

Autonomous navigation is a cornerstone of modern marine robotics, enabling ROVs to navigate complex underwater environments without direct human control \citep{lee2022autonomous}. Existing navigation systems often rely on pre-programmed routes or basic obstacle avoidance, limiting their adaptability. By integrating AI, particularly Large Language Models (LLMs) \citep{achiam2023gpt}, autonomous systems can interpret high-level commands and dynamically adjust their behavior based on real-time feedback.

Recent advancements in AI and LLMs have opened new avenues for enhancing autonomous systems' capabilities. LLMs have demonstrated remarkable performance in interpreting human commands, generating plans, and adapting to dynamic environments \citep{wang2024survey}. For example, Yang et al. \citep{yang2023oceanchat} introduced an innovative framework that integrates LLMs with robotics, allowing Autonomous Underwater Vehicles (AUVs) to interact seamlessly with human operators. Similarly, OceanPlan \citep{yang2024oceanplan} proposed a hierarchical LLM-driven planning and replanning framework, translating human instructions into actionable AUV controls to ensure reliable performance in complex underwater scenarios. Additionally, Word2Wave \citep{chen2024word2wave} developed a language-based interface for dynamic programming of AUVs, further advancing the integration of natural language processing in underwater robotic systems.


Despite these advancements, existing LLM-based systems often overlook the specific challenges of aquaculture net pen inspection, such as real-time adaptive navigation and precise inspection in dynamic underwater conditions. While tools like OceanChat, OceanPlan, and Word2Wave address broader marine applications, they lack focus on the integration of language processing with low-level control for aquaculture tasks. This paper explores the integration of LLMs into the control architecture of ROVs for aquaculture net pen inspection, introducing AquaChat- an LLM-guided navigation framework designed to improve inspection efficiency and accuracy. By translating high-level user commands into executable ROV tasks, AquaChat enables more intuitive and adaptive control, facilitating seamless communication between operators and autonomous systems.

The proposed framework employs a multi-layered architecture that includes an LLM-based planning layer, a task management layer, and a low-level control layer. A general overview of the framework is shown in Figure~\ref{fig:workflow}. Our approach employs LLMs to generate structured sequences of actions tailored to efficiently accomplish aquaculture net pen inspection tasks. For instance, consider a scenario where an ROV is asked to inspect multiple sections of a net pen to identify holes or areas of biofouling. The LLM dynamically computes a symbolic plan that optimizes the inspection path, prioritizing critical sections and minimizing overall mission time. If an unexpected condition, such as an obstruction or low visibility, prevents the ROV from completing a planned maneuver, the low-level control system provides real-time feedback to the LLM. This feedback allows the LLM to adapt the symbolic plan, addressing the issue and ensuring robust and precise inspection, even in dynamic underwater environments. By using LLMs' contextual understanding and planning capabilities, AquaChat enhances ROV autonomy, enabling more precise navigation and inspection capabilities within aquaculture environment. 

The key contributions of this work are;
\begin{itemize}
    \item Developing an LLM-guided framework for autonomous net pen inspection.
    \item Designing a multi-layered architecture that integrates planning, task management, and low-level control.
    \item Demonstrating the system’s effectiveness in simulated aquaculture environments, showcasing improvements in navigation accuracy and inspection capability.
\end{itemize}

The remainder of this paper is organized as follows: Section~\ref{sec:rw} reviews related work in aquaculture automation, net pen inspection, and marine robotics. Section~\ref{sec:problem} outlines the problem formulation. The proposed framework is detailed in Section~\ref{sec:framework}. Section~\ref{sec:sim-dep} discusses the simulation setup and deployment strategies. Section~\ref{sec:res} presents the experimental results and performance evaluation. Finally, Section~\ref{sec:conc} concludes the paper with key insights and potential future applications.

\section{Related Work}\label{sec:rw}
In this section, we review the current literature on underwater robotics for aquaculture applications. The related work is categorized into three key areas: control systems, computer vision, and the use of LLMs for inspection, navigation, and control.

\subsection{ROV Control system}
Control systems form the backbone of ROVs used in aquaculture, enabling precise and reliable navigation within challenging underwater environments \citep{akram2022robust}. Given the dynamic nature of aquaculture sites—characterized by varying currents, obstacles, and the presence of marine life—robust control mechanisms are essential for ensuring mission success. Advances in adaptive control schemes, robust path planning, and real-time feedback systems have significantly improved the operational efficiency of ROVs, making them well-suited for complex tasks like net pen inspection and maintenance.

Recently, significant efforts have focused on developing advanced control schemes for ROV navigation in aquaculture environments. For instance, Nguyen et al. \citep{nguyen2024robust} proposed two adaptive nonlinear controllers designed for robust velocity and heading control, addressing different maneuvering scenarios. Stability proofs and experimental results demonstrated their effectiveness, even in the presence of substantial environmental disturbances. Ohrem et al. \citep{ohrem2022robust} proposed a nonlinear dynamic positioning controller for ROVs in aquaculture environments, ensuring robustness against model uncertainties and disturbances. Stability analysis and field trials demonstrate its effectiveness, with simulations and real-world tests confirming reliable performance. Another adaptive controller is proposed in \citep{ohrem2023adaptive,kelasidi2022autonomous} by considering surge and sway speed under unknown paramters and external disturbances. In \citep{ohrem2024application}, a modified model reference adaptive controller and observer with an integral term to control surge and sway speed of an ROV is proposed. 

Botta et al. \citep{bott2024} developed an integrated system for ROV localization and mapping making use of acoustic and camera sensors, performing the net distance estimation and 3D map generation in real-time aquaculture inspection. Xia et al. \citep{xia2023scale}  proposed a scale-aware monocular odometry method using the Direct Sparse Odometry (DSO) technique for accurate position estimation in fishnet inspections. A region-of-interest extraction method and mesh alignment to the fishnet model allow the system to handle repeated textures and weak features, with successful evaluation in sea trials. In \citep{bjerkeng2023absolute, bjerkeng2021rov}, ROV localization method is proposed making use of onboard compass and laser-camera triangulation method. Cardailac et al. \citep{cardaillac2023application, cardaillac2024rov} proposed an approach that combines online path planning and following using multibeam forward-looking sonar and create an inspection map of the covered areas. 

In \citep{skaldebo2024approaches, schellewald2021vision}, the authors investigated the use of Fourier transformation with camera images for ROV relative pose estimation, stereo vision for depth map generation, and scanning sonar for pose estimation. A hybrid control scheme is introduced in \citep{wu2022intelligent} by integrating the neural network with proportional integral differential (PID) for net inspection. Amundsen et al. \citep{amundsen2021autonomous} proposed solution for autonomously guiding an ROV through an aquaculture net pen using Doppler velocity log (DVL) measurements to approximate the net's geometry. The approach minimizes cross-track error using a nonlinear guidance law, with simulations and experiments demonstrating its effectiveness, even in the presence of ocean currents. Rosa et al. \citep{rosa2024forward} proposed the use of forward-looking sonar method for ROV relative pose estimation for aquaculture net pens inspection. Similarly, the study \citep{tun2023development} explored various PID variants for aquaculture net inspection incorporating external disturbances. In \citep{lee2022autonomous}, another approached is developed that apply deep learning method on camera images for predicting ROV relative distance to the net. Similarly, Akram et al. \citep{akram2022visual} proposed vision-based net-distance estimation using the lines estimation on the net surface, and control law to for distance keeping during the operation. 

Research on hardware and system integration is also critical for ensuring that ROVs can handle the unique challenges of aquaculture environments. Some studies also focused on the system development and integration for aquaculture inspection tasks. For instance, Bent at al. \citep{haugalokken2024low} integrated various low-cost sensing modules into a standard BlueROV2, enhancing its capabilities for real-time aquaculture net inspection. The study \citep{rahim2022design} developed a customized ROV for water quality assessment in aquaculture environment. The study \citep{tarwadi2020design} also developed a compact, cost-effective ROV prototype for shallow-water inspections. The ROV is equipped with onboard real-time processing, camera-based piloting, and navigation sensors, and has been tested in a pool environment. The study \citep{vasileiou2021low} presented the design, fabrication, and control of a modular, affordable ROV for shallow-water inspections. Equipped with six thrusters and a 3D-printed hull, it offers customizability, increased maneuverability, and positive buoyancy, making it suitable for reliable operation in dynamic environments.

\subsection{Computer Vision for Aquaculture}

Computer vision has become a critical tool in automating underwater inspection tasks, particularly in the detection of defects such as net holes and biofouling in aquaculture net pens \citep{xu2023vision,aquadata}. Deep learning-based models, such as convolutional neural networks (CNNs), are extensively used for net defect detection and segmentation, enabling more accurate and efficient identification of anomalies compared to traditional inspection methods \citep{akram2024aquaculture}.

Numerous studies have explored the use of computer vision and deep learning techniques for aquaculture net pen inspection. For instance, Akram et al. \citep{akram2023evaluating} conducted a comparative analysis of various YOLO-based models for detecting net holes in aquaculture environments. In \citep{akram2023autonomous}, a deep learning model was integrated with an ROV to perform real-time net inspection tasks. Madshaven et al. \citep{madshaven2022hole, schellewald2022irregularity} proposed using the U-Net segmentation model \citep{unet} for detecting net holes using video streaming from ROV cameras in aquaculture settings. In \citep{zacheilas2021fpga,paspalakis2020automated}, a traditional vision-based method was introduced to detect net holes in real-time video input, accounting for lighting variations and haze. A two-method approach combining traditional vision techniques and YOLOv5 \citep{yolov5} was developed in \citep{paraskevas2023detecting,paraskevas2022detecting} for detecting net holes.

Further advancements include a mesh-hole grouping algorithm presented by Kang et al. \citep{kang2023detection} to detect damaged areas in fish cage nets by analyzing the differences in neighboring hole sizes from binary underwater images. Qiu et al. \citep{qiu2020fishing} proposed an automated framework for estimating the health of net pens by analyzing fouling and estimating the percentage of blockage from ROV images. Zhang et al. \citep{zhang2022netting} improved the Mask R-CNN model for detecting and locating marine netting damage, integrating Recursive Feature Pyramid (RFP) and Deformable Convolution Network (DCN) structures. This enhanced method achieved a 94.48\% average precision and improved detection speed, providing a robust solution for marine aquaculture environments.

In \citep{lopez2023inspection}, an integrated platform was developed, combining computer vision with a CNN to predict the distance between the net and ROV, along with an object detection algorithm to identify net holes. A simulation environment was also created to assess inspection trajectory algorithms. Betancourt et al. \citep{betancourt2020integrated} proposed an ROV-based solution for monitoring water quality and inspecting aquaculture net pens using real-time video and integrated sensors. Their system demonstrated 91\% accuracy in net pattern reconstruction and 79\% accuracy in net hole detection under various underwater conditions.Akram et al. \citep{akram2024aquaculture} proposed a novel multi-scale semantic segmentation model for aquaculture net pen inspection, focusing on detecting net holes, vegetation, and biofouling anomalies. The model was rigorously tested on a custom-developed net pen dataset, achieving over 90\% detection accuracy under various underwater conditions.

These studies highlight the growing role of computer vision and deep learning in transforming aquaculture net pen inspection, offering more precise and scalable solutions for real-time monitoring and damage detection.

\subsection{LLMs for marine Robotics}
The emergence of LLMs has revolutionized natural language processing (NLP) and is now extending its impact into robotics and control systems \citep{wang2024survey}. LLMs, such as GPT and similar models, can interpret complex commands, contextualize data, and generate adaptive plans for diverse tasks. In the field of robotics, LLMs have shown potential in generating symbolic plans, interpreting sensor data, and facilitating human-robot interactions \citep{achiam2023gpt}. By integrating LLMs with traditional control systems, researchers have developed frameworks where LLMs translate high-level mission objectives into actionable plans \citep{yang2023oceanchat}. These capabilities are particularly useful in dynamic environments like aquaculture, where real-time decision-making is essential \citep{yang2024oceanplan}.

In marine research, there is also new trend of using LLM with the combination of robotics navigation and control. For instance, Yang et al. \citep{yang2023oceanchat} proposed a novel system integrating LLMs with robotics to enable Autonomous Underwater Vehicles (AUVs) to interact intuitively with humans. OceanChat employs an LLM-guided closed-loop task and motion planning framework, translating human commands into actionable plans while adapting to dynamic underwater environments with event-triggered replanning. A photo-realistic simulation platform, HoloRvo, verifies the system’s effectiveness in enhancing mission success rates and computational efficiency.

OceanPlan \citep{yang2024oceanplan} is proposed contains a hierarchical LLM-based planning and replanning framework that translates human commands into actionable AUV control, ensuring robust operation in dynamic underwater environments. By decomposing long missions into manageable subtasks and integrating real-time feedback from a holistic replanner, our system addresses marine robotics challenges. Experimental results demonstrate improved planning efficiency and reliable AUV performance over extended missions using natural language input.

OCEANGPT \citep{bi2023oceangpt} is developed for specializing in ocean science tasks. This offers a DOINSTRUCT framework generates ocean-domain instruction data through multi-agent collaboration, and we present OCEANBENCH, the first oceanography benchmark for evaluating LLMs. Experiments show that OCEANGPT excels in ocean science knowledge and demonstrates emerging embodied intelligence capabilities in ocean technology.

Word2Wave \citep{chen2024word2wave} presented a language-based interface for dynamic programming of AUVs. W2W features novel language rules, a GPT-based prompt engineering module, and an SLM (T5-Small)-based learning pipeline for converting human speech/text into AUV missions. Benchmark evaluations and user studies highlight its superior performance compared to commercial interfaces.

In \citep{samuel2024integrating}, a real-time, vision-based surveillance system is proposed using LLMs and a Raspberry Pi monitors surface-level water pollution, detecting seven major pollutants like algal blooms and oil spills. It generates contextual insights about pollution types, causes, and impacts, helping authorities respond proactively. The system autonomously alerts local authorities, reducing the need for human intervention.

FathomGPT \citep{khanal2024fathomgpt} is an open-source tool designed for interactive exploration of ocean science data using natural language. Developed with marine scientists, it enables advanced queries, image retrieval, taxonomic mapping, and on-demand chart generation from the Fathom-Net database. Emphasizing user experience, the system uses OpenAI’s language models, with studies showcasing its effectiveness and feedback from experts.

Another study \citep{lian2024evaluation} evaluates SAM2's potential in marine science using the UIIS and USIS10K underwater segmentation benchmarks. SAM2 performs excellently with ground truth bounding box prompts but struggles in automatic mode with point prompts. The findings aim to encourage further research on SAM models for underwater applications.

While significant progress has been made in using LLMs for marine robotics, notable gaps still exist in the current state of research. Existing systems such as OceanChat, OceanPlan, and Word2Wave emphasize task and motion planning, hierarchical frameworks, and language-based programming for AUVs. However, these approaches primarily target broad marine applications, often overlooking domain-specific challenges like aquaculture net pen inspection. Additionally, tools like FathomGPT and OCEANGPT excel in ocean data exploration and domain-specific knowledge generation but do not directly address the complexity of real-time navigation, precise inspection, and adaptive control required for aquaculture environments.  

Our work bridges these gaps by presenting AquaChat, a novel LLM-guided framework explicitly tailored for ROV-based aquaculture net pen inspection. Unlike prior studies, AquaChat focuses on integrating high-level natural language understanding with low-level feedback loops for adaptive navigation and precise inspection in dynamic underwater settings. This work not only advances the state-of-the-art in applying LLMs to marine robotics but also contributes a specialized solution addressing the unique demands of aquaculture operations.

\section{Problem Formulation}\label{sec:problem}
In this section, we illustrate and formulate the aquaculture net pens inspection problem. 

The ROV receives real-world observation $z_t$ from its onboard sensors and camera, capturing the environment state and record data. A human command $q$ represents an abstract textual input, describing inspection tasks for the ROV. Planning directly from the high-level command $q$ to low-level control $u_t$ is challenging due to the long-horizon nature of the problem. Therefore, we decompose this overall task into three sub-problems. 

\begin{itemize}
    \item Problem 1 (Plan generation): Translate the human command $q$ into a concrete inspection plan $\mathcal{P}$, consisting of a sequence of symbolic actions \hbox{$\mathcal{P}=\{a_0,a_1,\ldots , a_n\}$}.
    
    \item Problem 2 (Task mapping): Map each symbolic action $a_i$ into executable ROV actions, ensuring preconditions are satisfied.

    \item Problem 3 (Action execution): Control the ROV to perform actions such as movement, inspection and recording in a feedback-driven manner, using real-time observations $z_t$ for error correction and adaptive control. 
\end{itemize}

In the 3D underwater environment, we define the ROV's state at time $t$ as $s_t =[x,y,z,\alpha,\beta,\gamma]^T$ $\in$ $S$, where, $S$ is the state space contains all valid states of the ROV, $(x,y,z)$ represents the position in Cartesian coordinates, $(\alpha,\beta,\gamma)$ are roll, pitch, and yaw angles represents the orientation of the ROV. The control input $u_t$ is defined as $u_t=[\upsilon_t, \omega_t]^T$ where, $\upsilon_t$ and ,$\omega_t$ are linear and angular velocities respectively. 

Let LLM act as a block box that takes the task instruction in term of text as input and generate a sequence of actions $\mathcal{P} = \{ a_1, a_2, \dots, a_n \}, \quad a_i \in \{ \text{move\_to}, \text{inspect}, \text{capture} \}$, each action $\alpha_i$ corresponds to a specific task as follows:

The task \textit{move\_to} is defined to navigate ROV to the goal state $(x_g, y_g, z_g,\alpha_g,\beta_g,\gamma_g)$, 
\textit{move\_to}\((x_g, y_g, z_g,\alpha_g,\beta_g,\gamma_g) \Rightarrow u_t \text{ such that } \lim_{t \to \max} s_t =  [x_g, y_g, z_g,\alpha_g,\beta_g,\gamma_g]^T \)

The task \textit{Inspect} define the inspection trajectory. The ROV moves in a helical path around the net pen, parameterized as:
\[
x(t) = r \cos(\omega t), \quad y(t) = r \sin(\omega t), \quad z(t) = z_0 - v_z t
\]

The task \textit{capture} perform data recording using the camera input $I_t$ at positions $s_t$.

Our LLM-guided aquaculture inspection framework is defined as the tuple $\langle P, K \rangle$, where prompt $P$ encodes the task description, initial state, and inspection goals, including: user instruction $q$: ``Inspection the net pens for defects'', initial state: $s_o$, environment: Net pen dimensions, ROV specs, and current observation. $K$ contains the set of predefined constraints about inspection protocols such as avoiding certain regions while inspection.The output of $a$ is an ordered sequence of actions: $a=\{\text{move\_to},\text{inspect},\text{capture}\}$. 





This formal framework ensures a structured and adaptive approach to underwater net inspection using LLM guidance.

\begin{figure*}[t]
    \centering
    \includegraphics[width=1\linewidth]{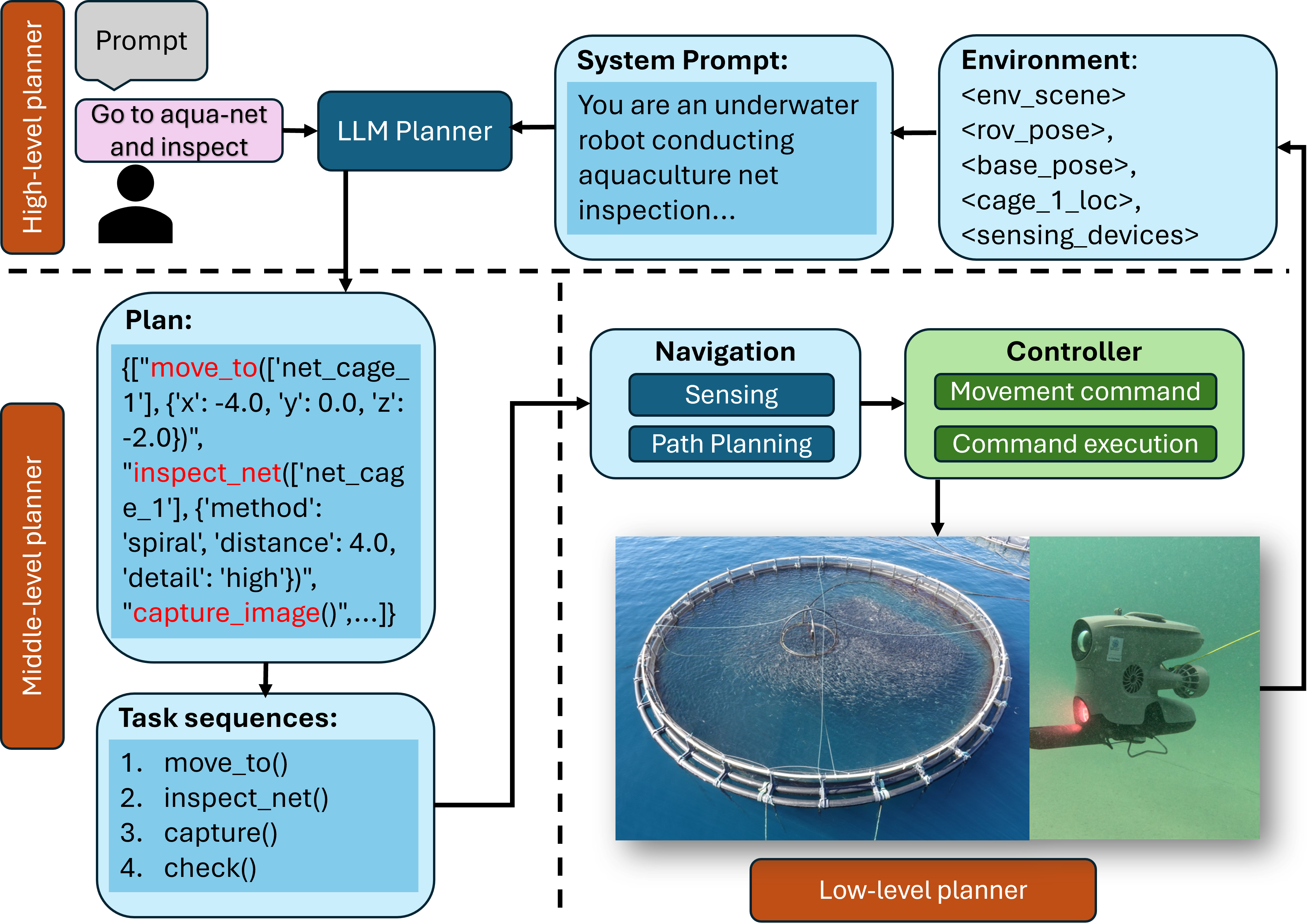}
     \caption{Proposed framework for LLM-guided navigation in aquaculture net pen inspection using ROV. The system comprises three modules: (1) High-level planner: Translates user commands into a symbolic inspection plan using GPT-4; (2) Middle-level planner: Generates a sequence of actions based on the inspection plan; (3) Low-level planner: Executes task sequences adaptively by integrating sensing, path planning, and control modules. Real-time observations provide feedback and enhance perception throughout the inspection process.}
    \label{fig:framework}
\end{figure*}

\section{Proposed Framework}\label{sec:framework}
\subsection{Overview}
The primary objective of this research is to develop a real-time, LLM-guided inspection framework that translates high-level natural language commands into detailed ROV navigation and inspection tasks. The system ensures robust execution through a feedback mechanism, enabling the ROV to dynamically adapt to environmental conditions and task progress.

The proposed framework is show in Figure~\ref{fig:framework}. To execute ROV inspection missions defined by human commands, we propose a three-tier framework for AquaChat, which integrates closed-loop LLM-guided task and motion planning. This framework operates in three stages: first, the LLM interprets the human command; second, the task planner sequences the tasks; and third, the motion planner controls the ROV within the aquaculture environment.

The inspection plan begins with the ROV navigating to a base location. From this point, it orients itself towards the center of the net pen and follows a systematic, top-to-bottom helical trajectory to ensure full coverage of the net structure. During the inspection phase, the camera continuously captures images, that are useful for further inspection and analysis to asses the current state of the net. 

By integrating natural language processing, adaptive planning, navigation and control for net pens inspection, AquaChat offers a novel approach to enhancing efficiency, accuracy, and scalability in aquaculture environment.

\subsection{LLM-Based Planner}

The LLM-Based Planner constitutes the core of the proposed framework, facilitating the translation of abstract user commands into actionable robotic missions. By using LLM, the planner ensures seamless interpretation, decomposition, and contextual adaptation of tasks for robust aquaculture net pen inspection. This section details the components and functionalities of the LLM-Based Planner.

\begin{figure}
    \centering
    \includegraphics[width=1\linewidth]{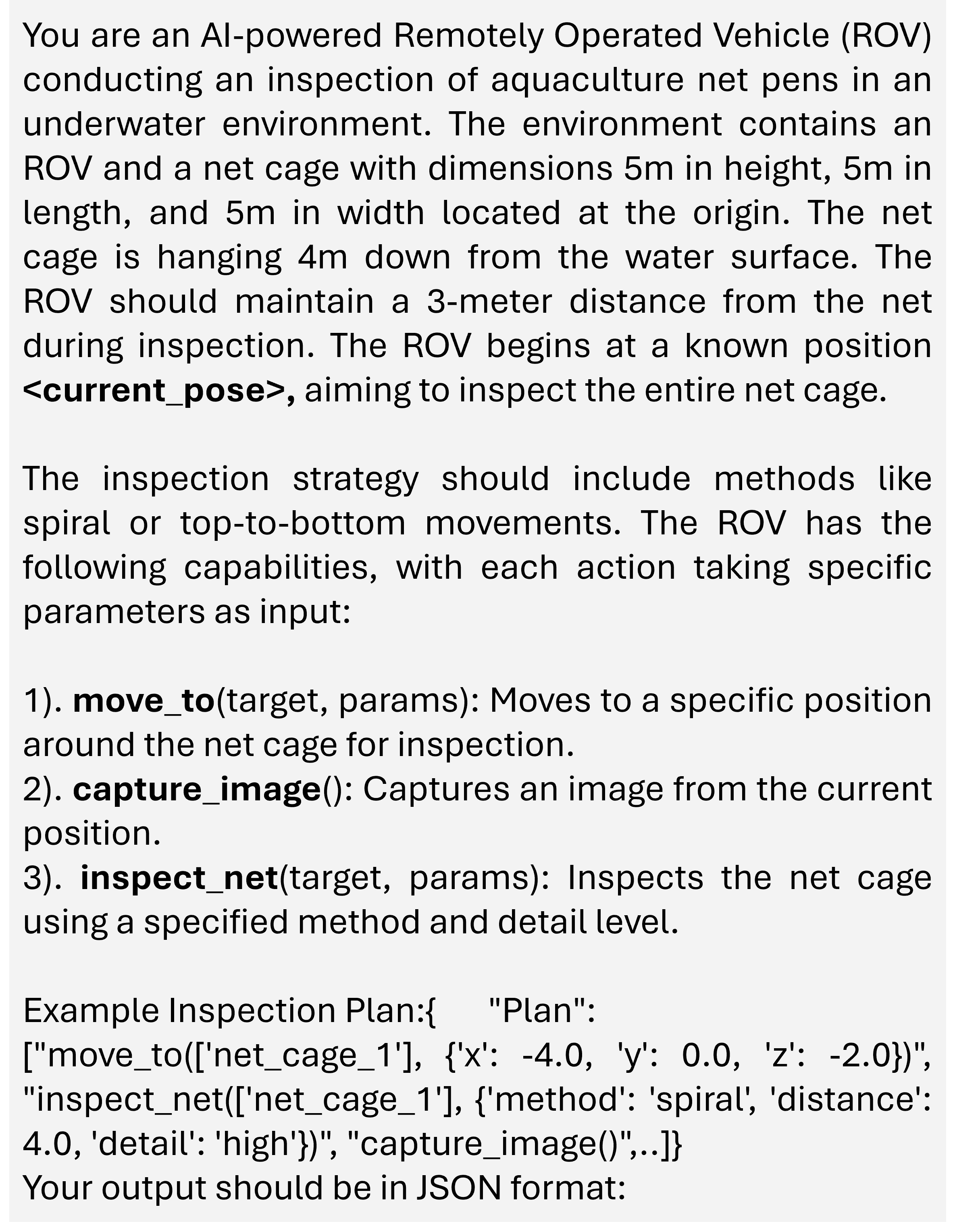}
    \caption{Example of prompt used for aquaculture net pens inspection. This prompt integrates the environment description and the system specification along with the plan constrains.}
    \label{fig:prompt}
\end{figure}

The main objective of the LLM-Based Planner is to process high-level user commands, such as ``Inspect the aquaculture net pen for defects''. The LLM parses these commands, integrating them with predefined contextual information about the environment and the ROV's operational specifications. This contextual data encompasses the dimensions of the net pen, the ROV's sensory and mobility capabilities, and other mission-critical parameters. The template of the final prompt is shown in Figure\ref{fig:prompt}. The output is a symbolic plan that decomposes the high-level instruction into discrete, manageable tasks aligned with the mission goals. For example, a command to inspect a net pen may be decomposed into tasks such as navigate to the net pen, position for inspection, capture images, and detect defects etc.

The LLM translates the decomposed tasks into a sequence of symbolic actions, each corresponding to an operational goal. This symbolic plan adheres to the constraints of the operational environment and the ROV’s specifications. For instance, the generated plan considers factors such as the net pen’s spatial configuration, the ROV's range of motion, and the requirements for capturing high-quality images. The LLM ensures the sequence aligns with the inspection workflow, typically consisting of actions like move\_to, inspection, capture, and defect detection. By encoding these tasks into symbolic representations, the planner bridges the gap between abstract commands and executable operations.

\begin{figure}[h]
    \centering
    \includegraphics[width=1\linewidth]{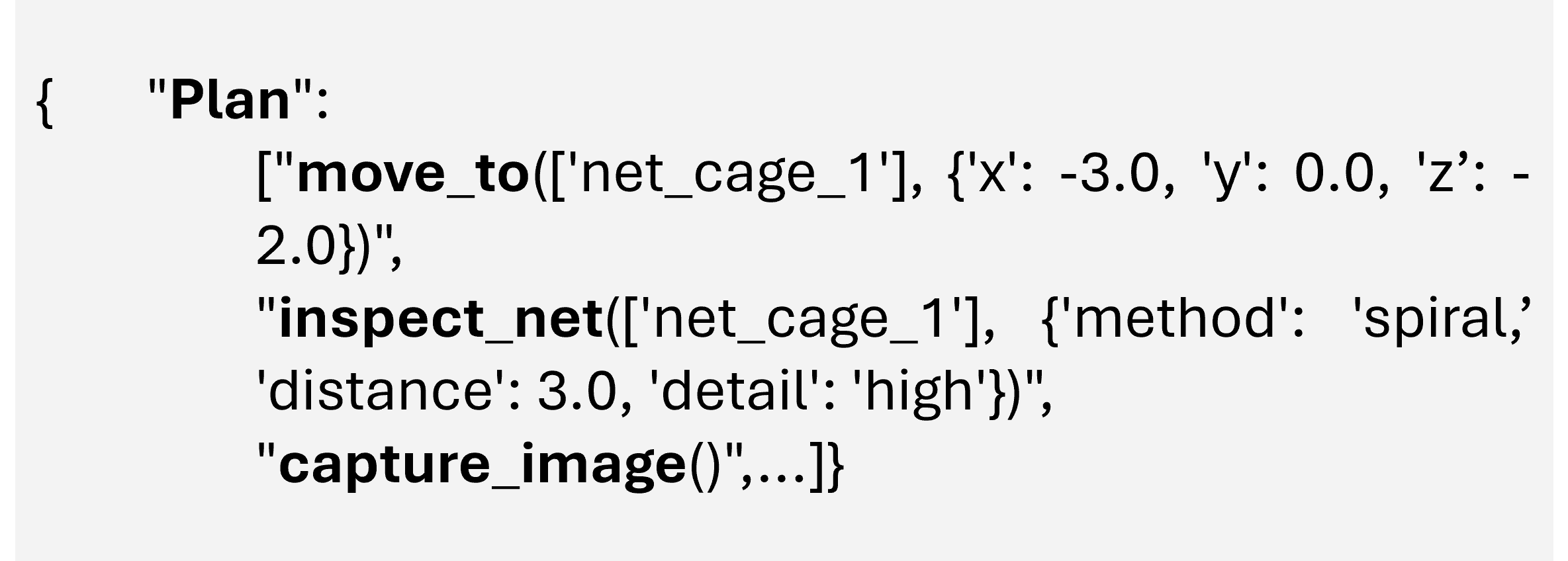}
    \caption{Output of the LLM Planner for aquaculture net pens inspection based on the user input. The plan consists of a sequences of actions to be executed by the ROV.}
    \label{fig:full-plan}
\end{figure}

Given a high-level user command $q$, the LLM parses in into a set of symbolic tasks  $\mathcal{P} = \{ a_1, a_2, \dots, a_n \}$ where each task $a_i$ represents a discrete action such as navigation, inspection, or image capture. 
The user commands is also integrated with the contextual information about the environment, such as net pen dimensions, location, boundaries, and operational constraints as well as the ROV's specification, including sensory data and properties. For example, a user prompt c=``Inspect the aquaculture net pen'' might be decomposed into plan $\mathcal{P}$=\{move\_to, inspect, capture\}. Then, the planner generate a symbolic plan $P$ as a sequence of actions $P=\{a_1,a_2, \dots , a_n\}$ with each action $a_i$ formulated based on environmental constraints and the mission goal. The generation of the plan can be expressed as:

The LLM Planner generates a symbolic plan based on the user input. For example, a user input ``Go to cage-1 and inspect the entire net cage using a spiral method at a 3-meter distance, and take a photo''. Figure~\ref{fig:full-plan} shows the output of the LLM planner. This plan outlines the sequence of steps required to complete the mission, such as moving to a base station, perform inspection, and capturing a photograph. By focusing on what to do instead of how to do it, the planner creates an adaptable roadmap to achieve the objectives of the mission.

One of the key strengths of the LLM-Based Planner is its ability to dynamically adapt the symbolic plan based on real-time feedback and environmental inputs. During the inspection process, the ROV continuously updates the planner with observations from onboard sensors. For instance, if unexpected obstacles or changes in the net pen’s structure are detected, the LLM reassesses the plan, reconfiguring tasks to accommodate the new context. This dynamic adaptation mechanism ensures robust mission execution, reducing the likelihood of failure due to unforeseen environmental factors.

By integrating high-level user input with contextual knowledge and real-time feedback, the LLM-Based Planner serves as an intelligent intermediary between human operators and the ROV. This capability not only enhances the efficiency of aquaculture net pen inspections but also highlight the potential of LLMs in advancing robotic autonomy in complex marine environments.

\subsection{Mid-Level Task Planner}

The Low-Level Planner bridges the gap between the high-level symbolic plan generated by the LLM-Based Planner and the actionable commands executed by the ROV. This layer ensures precise execution of tasks while dynamically responding to environmental uncertainties and system feedback.

In robotic systems like ROV-based aquaculture net pen inspection, the absence of a middle-level planner poses significant challenges in translating high-level plans into actionable commands. Without this intermediary layer, the system struggles to break down complex symbolic plans into granular low-level actions, resulting in misinterpretations and execution failures. Moreover, the lack of precondition validation means that tasks may proceed under invalid conditions, leading to incomplete or erroneous operations. Error handling is also limited, as the system cannot dynamically adapt to real-time feedback or environmental changes, causing abrupt mission failures in the event of unforeseen obstacles or system malfunctions. As mission complexity increases, directly managing high-level tasks at the low-level layer becomes unsustainable, creating scalability issues and complicating the debugging process.

In contrast, a middle-level planner introduces modularity, adaptability, and robust error handling, bridging the gap between high-level planning and low-level execution. By decomposing symbolic plans into precise actions, it ensures systematic execution while validating preconditions before task initiation. The middle layer also facilitates dynamic adaptation, monitoring for errors or environmental changes, and triggering replanning when necessary. This capability not only enhances error recovery but also maintains mission continuity without requiring high-level intervention. Additionally, the modular structure of the middle-level planner simplifies the integration of high-level strategies with low-level operational constraints, making the system scalable and efficient for complex tasks. Overall, the middle-level planner is essential for achieving reliable, adaptive, and cohesive performance in autonomous robotic missions.

In autonomous robotic systems, defining clear Boolean-valued predicates and mapping them to corresponding actions is essential for robust and efficient task execution. The absence of a middle-level planner leads to significant limitations in managing these relationships, particularly in tasks such as aquaculture net pen inspection using ROVs. Below, we highlight the logical framework enabled by a middle-level planner and the associated challenges without it.

We define the following Boolean-valued predicates to represent the state of the ROV and its environment:

\begin{itemize}
    \item navigated(rov - ROV): Evaluates to True if the ROV has successfully navigated to the designated position.
     \item inspected(area - Region): True if a specific area has been inspected by the ROV.
     \item captured(area - Region): True if images of the inspected area have been successfully captured.
     \item position\_valid(rov - ROV): True if the ROV’s position matches the target coordinates.
     \item system\_ready(): True if the ROV’s systems are operational and prepared for task execution.
     \item replanned(): True if a replanning signal is sent to adjust the task sequence
\end{itemize}

The ROV actions are defined as follows, corresponding to the predicates:

\begin{itemize}
    \item navigate(rov): Direct the ROV to a specific position.
    \item inspect(area): Inspect a designated region of the net pen.
    \item capture(area): Capture images of the inspected region.
    \item report(area): Send the inspection results, including detected defects and captured images, to the operator.
    \item replan(): Trigger the replanning process when preconditions are unmet or unexpected events occur.
\end{itemize}

In the following, we present detailed preconditions and effects of each action associated with the overall framework. First, a domain definition \texttt{aquachat\_inspection} is established, specifying the requirements for mission planning and object typing. The \texttt{:types} defines two primary object classes, ROV and Environment, ensuring structured modeling of the system's components.

\begin{lstlisting}[language=PDDL,caption={}]
(define (domain aquachat_inspection)
  (:requirements :strips :typing)
\end{lstlisting}

\begin{lstlisting}[language=PDDL,caption={}]
  ;; Types
  (:types ROV Environment)
\end{lstlisting}

The predicates defines the key conditions that represent the state of the system and its components in the AquaChat framework. These predicates are used to evaluate various aspects of the ROV's operation and the environmental conditions during the inspection process. For example, \texttt{`system\_ready`} indicates whether the ROV's systems are operational, while \texttt{`region\_detected`} and \texttt{`environment\_stable`} verify the readiness of specific areas and the overall environment for inspection tasks. The \texttt{`move\_to`} predicate checks if the ROV has reached the designated position, and \texttt{`inspected`} and \texttt{`captured`} track whether a region has been inspected and images have been captured. Additionally, planning-related predicates like \texttt{`plan\_ready`} and \texttt{`plan\_executed`} manage the task sequence, while \texttt{`feedback\_received`}, \texttt{`replan\_needed`}, and \texttt{`replan\_completed`} handle the feedback and replanning processes. These predicates collectively provide the necessary conditions for managing the ROV's tasks and ensuring the efficiency of the inspection process.

\begin{lstlisting}[language=PDDL,caption={}]
  ;; Predicates
  (:predicates
    (system_ready)
    (region_detected ?area - Region)
    (environment_stable)
    (move_to ?rov - ROV)
    (inspected ?area - Region)
    (captured ?area - Region)
    (defect_detected ?area - Region)
    (plan_ready)
    (plan_executed)
    (feedback_received)
    (replan_needed)
    (replan_completed)
  )
\end{lstlisting}

The \textit{Plan} action triggers the initiation of the planning process, which is provided by the LLM planner module, once the system is ready and planning has not been completed. Its effect sets the \texttt{plan\_ready} predicate, signaling that the system is prepared to execute the tasks.
\begin{lstlisting}[language=PDDL,caption={}]
  ;; Plan
  (:action plan
    :parameters ()
    :precondition (and (system_ready) (not (plan_ready)))
    :effect (plan_ready)
  )
\end{lstlisting}

The \textit{Move-to} action directs the ROV to a specified area within the environment. It takes two parameters: the ROV \texttt{(`?rov`)} and the target region \texttt{(`?area`)}. The action's precondition ensures that the planning process is ready \texttt{(`plan\_ready`)}, the target region has been detected \texttt{(`region\_detected`)}, and the environment is stable \texttt{(`environment\_stable`)}. Once these conditions are met, the action’s effect is to mark the ROV as having successfully navigated to the target area \textit{Move-to}. This action is essential for enabling the ROV to move autonomously to specific inspection locations.

\begin{lstlisting}[language=PDDL,caption={}]
  ;; Move-to
  (:action move_to
    :parameters (?rov - ROV ?area - Region)
    :precondition (and (plan_ready) (region_detected ?area) (environment_stable))
    :effect (navigated ?rov)
  )
\end{lstlisting}

The \textit{Inspect} action directs the ROV to inspect the aqua-net after it has navigated to the base location and followed a generated trajectory. Once the preconditions are met, the effect is that the area is marked as inspected \texttt{(`inspected ?area`)}.

\begin{lstlisting}[language=PDDL,caption={}]
 ;; Inspect
(:action inspect
  :parameters (?rov - ROV ?area - aqua-net)
  :precondition (and 
    (navigated ?rov) 
    (region_detected ?area)
    (trajectory_generated ?rov ?area)  
  )
  :effect (inspected ?area)  
)
\end{lstlisting}

The \textit{Capture} action ensures that the ROV is at the specified location and the area has been inspected before capturing images. Once the preconditions are met, the effect marks the area as captured \texttt{(`captured ?area`)}.

\begin{lstlisting}[language=PDDL,caption={}]
  ;; Capture
(:action capture
  :parameters (?rov - ROV ?area - aqua-net)
  :precondition (and 
    (inspected ?area) 
    (move_to ?rov)  
  )
  :effect (captured ?area)  
)
\end{lstlisting}

Similarly, \textit{Replan} and \textit{Report} actions are defined below. \textit{Replan} takes feedback from the ROV sensors and ensure if there is an error or failure in mission execution, it recall the plan. The \textit{Report} action is used to mark the ROV mission completion. 

\begin{lstlisting}[language=PDDL,caption={}]
  ;; Replan
  (:action replan
    :parameters ()
    :precondition (and (feedback_received) (replan_needed))
    :effect (replan_completed)
  )
\end{lstlisting}

\begin{lstlisting}[language=PDDL,caption={}]
  ;; Report
  (:action report
    :parameters (?area - aqua-net)
    :precondition (captured ?area)
    :effect (report_sent)
  )
\end{lstlisting}

Without a middle-level planner, these predicates and actions lack a cohesive framework for validation, sequencing, and adaptation. For instance, without evaluating the predicate \texttt{position\_valid(rov)}, the action \texttt{inspect(area)} might commence at an incorrect location, leading to incomplete or erroneous inspections. Additionally, if the predicate \texttt{system\_ready()} is not verified, the ROV might attempt tasks while its systems are non-functional, causing mission failure. Furthermore, in dynamic environments, the absence of adaptive replanning logic \texttt{(replanned())} means the system cannot respond to obstacles or errors, leading to mission termination or reduced reliability.
Each step is validated and executed based on its associated predicates, ensuring seamless transitions and error recovery. This logical framework highlights the indispensability of the middle-level planner in achieving reliable, adaptable, and efficient autonomous robotic missions within aquaculture environment.


\subsection{Low-Level Motion Planner}

The low-level motion planner is critical for the precise control and execution of actions in ROV-based aquaculture net pen inspection (see Figure~\ref{fig:workflow}). It translates high-level commands into actionable movement strategies, ensuring the ROV's navigation and inspection operations are executed accurately in a dynamic environment. This module uses the sensing and path planning capabilities. Sensors such as camera, GPS, and IMU provide real-time data about the environment, including obstacle locations and the position of the ROV and the installed aqua-net \citep{cahyadi2023performance}. Path-planning algorithms use these data to compute safe and efficient paths to the target, avoiding collisions and optimizing the travel path. 

The motion planner handles commands such as controlling position, velocity, orientation, and depth. A command like \texttt{move\_to(target\_position)} is processed by the low-level planner into the specific velocity and heading adjustments needed for the ROV to reach the target.

To begin the low-level motion planning, we first define the ROV's system model, which incorporates vehicle positions (\(x, y, z, \theta\)):

\begin{equation}\label{eq:rov_motion}
  \dot{x} = v \cos(\theta), \quad \dot{y} = v \sin(\theta), \quad \dot{z} = v_z, \quad \dot{\theta} = \omega  
\end{equation}

Where:
\begin{itemize}
    \item \( x, y, z \): Cartesian coordinates of the ROV.
    \item \( \theta \): Orientation angle (yaw) of the ROV.
    \item \( v \): Forward velocity (control input for horizontal movement).
    \item \( v_z \): Vertical velocity (control input for depth movement).
    \item \( \omega \): Turning rate (control input for yaw/rotation).
\end{itemize}
The system's state is represented as:

\begin{equation}
  x = \begin{bmatrix} x \\ y \\ z \\ \theta \end{bmatrix}  
\end{equation}

And the control input is:

\begin{equation}
  u = \begin{bmatrix} v \\ v_z \\ \omega \end{bmatrix}.  
\end{equation}

In a discrete-time model, the system dynamics are expressed as:

\begin{equation}
x_{k+1} = f(x_k, u_k),
\end{equation}

where \( k \) denotes the discrete time step. This formulation allows for numerical solutions to be computed for the motion planning tasks.

The ROV operates within regions of the aquaculture environment:

\begin{equation}
R = \{ r_1, r_2, \dots, r_M \},
\end{equation}

Each region \( r_i \) is associated with an environmental map \( V \), and the ROV’s position \( x_0 \) within a region \( r_i \) defines the planning problem. This motion planning problem can be formulated as an optimization problem:

\begin{equation}
\min_{u_0, \dots, u_N} J = \sum_{k=0}^{N} L(x_k, u_k),
\end{equation}

subject to:

\begin{equation}
x_{k+1} = f(x_k, u_k), \quad k = 0, 1, \dots, N-1,
\end{equation}

\[
x_0 \in r_i, V.
\]

Where:
\begin{itemize}
    \item \( J \) represents the cost function, typically minimizing time or energy for the motion task.
    \item \( L(x_k, u_k) \) is the stage cost at time step \( k \), considering the current state and control inputs.
    \item \( N \) is the number of steps in the planning horizon.
\end{itemize}

This framework captures the ROV's motion in required degrees of freedom, including horizontal and vertical movements, and allows for precise control of the inspection process in aquaculture environments.

Next, we present the general control formulation for two key actions of the ROV: \texttt{move\_to} and \texttt{inspect}. These formulations ensure that the ROV moves efficiently to a target location and inspects the net surface while maintaining smooth and stable control over its movements.

\subsubsection{Move-To Action Control}

The \texttt{move\_to} action involves navigating the ROV from its current position \(\mathbf{x}(t) = [x(t), y(t), z(t), \theta(t)]^\top\) to a desired target position \(\mathbf{x}_d = [x_d, y_d, z_d, \theta_d]^\top\), where \(\mathbf{x}(t)\) is the ROV's state at time \(t\) and \(\mathbf{x}_d\) is the desired position. The goal is to minimize the positional error between the current state and the desired state.

The objective is to minimize the error \(\mathbf{e}(t) = \mathbf{x}_d - \mathbf{x}(t)\), where:
\begin{equation}
    \mathbf{e}(t) = \begin{bmatrix} e_x(t) \\ e_y(t) \\ e_z(t) \\ e_\theta(t) \end{bmatrix} = \begin{bmatrix} x_d - x(t) \\ y_d - y(t) \\ z_d - z(t) \\ \theta_d - \theta(t) \end{bmatrix}
\end{equation}

The guidance law computes the desired control inputs \(\mathbf{u} = [v_x, v_y, v_z, \omega]^\top\) using a PID~\citep{ang2005pid} controller. The control laws for each component are as follows:
\begin{equation}\label{eq:control_law}
\begin{aligned}
 v_x = k_p \cdot e_x(t) + k_i \cdot \int_0^t e_x(\tau) \, d\tau + k_d \cdot \dot{e}_x(t),\\
v_y = k_p \cdot e_y(t) + k_i \cdot \int_0^t e_y(\tau) \, d\tau + k_d \cdot \dot{e}_y(t),\\
v_z = k_p \cdot e_z(t) + k_i \cdot \int_0^t e_z(\tau) \, d\tau + k_d \cdot \dot{e}_z(t),\\
\omega = k_p \cdot e_\theta(t) + k_i \cdot \int_0^t e_\theta(\tau) \, d\tau + k_d \cdot \dot{e}_\theta(t),   
\end{aligned}
\end{equation}

where \(k_p\), \(k_i\), and \(k_d\) are the proportional, integral, and derivative gains, respectively. These terms enable the ROV to adjust its velocity and heading based on the error dynamics, ensuring stability and smooth convergence to the target.

The ROV's motion is governed using Equation~\ref{eq:rov_motion}. The PID-based control law ensures that the computed control inputs \(\mathbf{u}\) guide the ROV to its target position by dynamically adjusting thrust, heading, and depth. This approach accounts for the ROV's dynamics and environmental disturbances, facilitating accurate execution of the \texttt{move\_to} and inspection actions in aquaculture net pen operations.

\subsubsection{Inspect Action Control}

The \texttt{inspect} action requires the ROV to navigate around a net surface and inspect it thoroughly. The task is to generate waypoints that ensure the ROV covers the entire net surface, which could be modeled as a series of waypoints based on the net's dimensions \(L, W, H\). Each waypoint \(\mathbf{x}_i = [x_i, y_i, \theta_i]\) represents the ROV's desired position and orientation.

Let the ROV follow a planned path around the net. We generate waypoints \(\{\mathbf{x}_1, \mathbf{x}_2, \dots, \mathbf{x}_n\}\) that describe a trajectory around the net. These waypoints can be generated using path-planning algorithms such as $A^*$ or a any heuristic approach based on the geometry of the net.

The objective during inspection is to guide the ROV along the path of waypoints while ensuring that it maintains a fixed distance from the net surface and adjusts its orientation to capture images or perform other inspection tasks. The error for each waypoint is defined as:
\begin{equation}
    \mathbf{e}_i(t) = \mathbf{x}_i - \mathbf{x}(t)
\end{equation}

where \(\mathbf{x}_i\) is the desired waypoint and \(\mathbf{x}(t)\) is the ROV’s current position.

A similar proportional control law (Equation~\ref{eq:control_law}) is used to guide the ROV between waypoints. The control inputs for each waypoint are calculated by adjusting the ROV's velocity and orientation. The ROV will execute these commands while adjusting its trajectory dynamically to avoid obstacles or deviations from the planned path. By employing these general control formulations, the ROV can navigate efficiently on the desired path and perform thorough inspections, while maintaining stability and robustness in its movement and control.

\section{Simulation and Deployment}\label{sec:sim-dep}


To evaluate the performance of AquaChat in guiding ROVs for aquaculture net pen inspection, we utilized a simulated environment designed in the ROS Gazebo framework. The simulation setup included the integration of the UUV Simulator~\citep{uuv_simulator}, BlueROV2 Simulator~\citep{bluerov2_simulator}, and a custom-designed aqua-net, which provided realistic underwater physics and control capabilities used for ROV operations.

The UUV Simulator is an advanced simulation framework designed for underwater vehicles, including ROVs and AUVs. Built on the ROS and Gazebo platforms, it provides a comprehensive environment for simulating underwater dynamics with high fidelity. Key features include realistic hydrodynamic modeling, accounting for drag, buoyancy, and added mass, which are critical for replicating underwater motion. The simulator also supports various sensors, such as sonar, depth sensors, and cameras, enabling the testing of perception and control algorithms. Additionally, the simulator includes specialized models, such as the ocean waves Gazebo model, which simulates realistic oceanic conditions, incorporating wave height, frequency, and direction. This model simulates the effects of dynamic ocean currents and wave-induced forces on the ROV, allowing for the testing of navigation and control systems under real-world conditions. The UUV Simulator is widely used in academic and industrial applications, making it an essential tool for the development and validation of underwater robotics systems \citep{akram2021visual}.

\begin{figure}[t]
    \centering
    \includegraphics[width=1\linewidth]{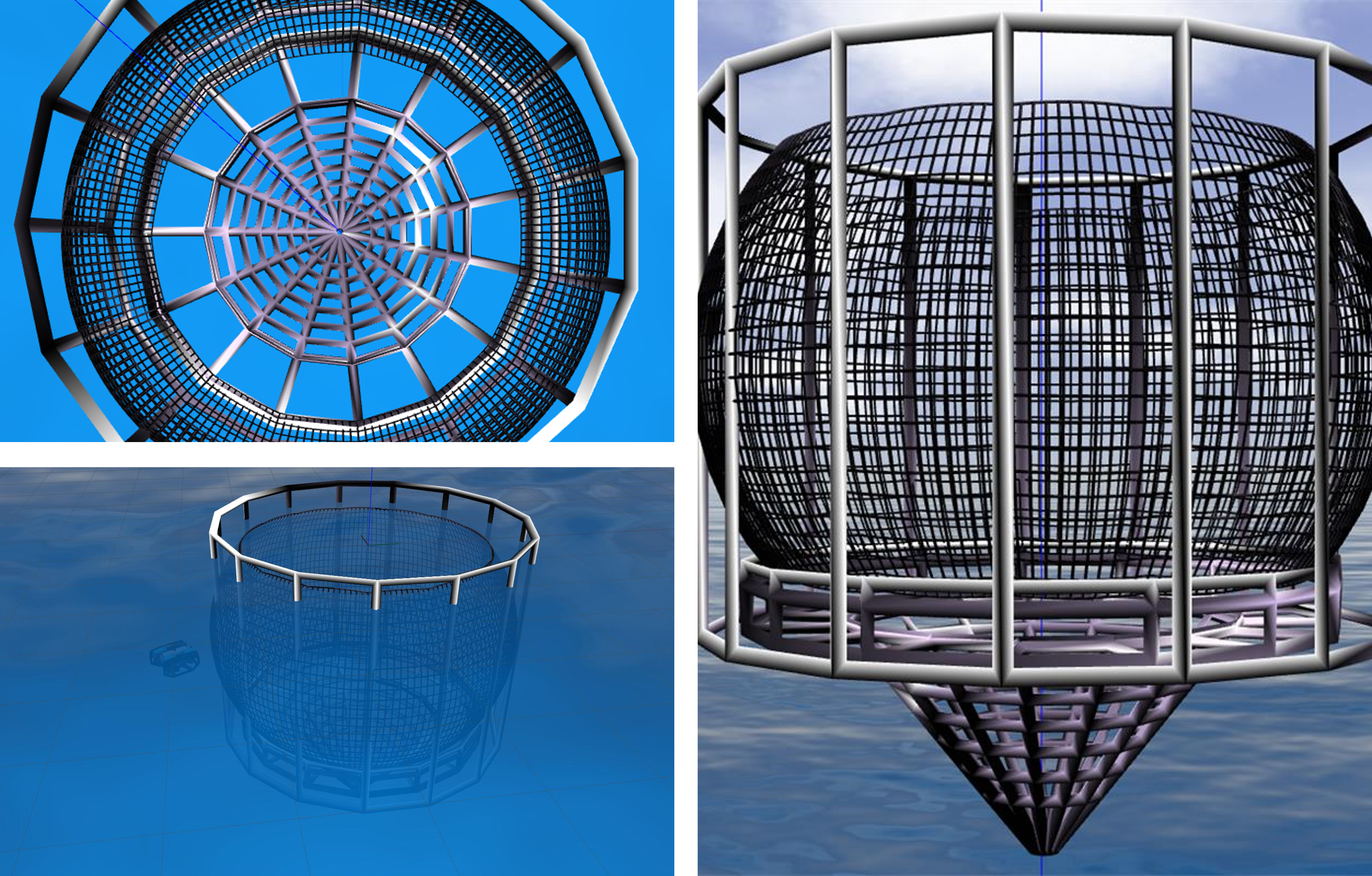}
    \caption{Aqua-net simulated 3D model: A detailed visualization of an aquaculture net pen featuring structural frames and mesh, designed for underwater inspection and maintenance using ROVs.}
    \label{fig:aqua-net}
\end{figure}

The BlueROV2 Simulator is a specialized simulation package designed to model the BlueROV2 \citep{BlueROV2}, one of the most widely used underwater vehicles for research and inspection tasks. Built within the UUV Simulator framework, it provides a detailed and accurate representation of the BlueROV2's physical characteristics, including its dimensions, thruster configuration, and mass properties. The simulator incorporates realistic underwater dynamics and allows for the integration of various sensors, such as cameras, sonars, and depth sensors, to replicate real-world scenarios. With support for ROS, it facilitates seamless communication with control algorithms, enabling researchers to test navigation, manipulation, and inspection tasks in a virtual environment. The BlueROV2 Simulator is an invaluable tool for developing and validating underwater robotic applications before deployment in real-world conditions \citep{von2022open}.

Additionally, a custom-designed aquaculture net pen model was developed to replicate the structural and environmental conditions typically encountered in fish farming systems \citep{akram2022visual}. The aquaculture net pen model was created using the Blender tool and further optimized to ensure compatibility with the ROS Gazebo environment. The model represents a spherical net pen structure, featuring:

\begin{itemize}
    \item Mesh Detailing: A realistic net mesh to mimic the structural patterns of actual aquaculture pens, enabling precise inspection tasks.
    \item Frame Design: Supporting beams and cylindrical components to simulate the structural integrity of the pen. The cage has a 5m height and 5m width, located at origin.
    \item Environmental Elements: The simulation environment incorporated water currents, lighting variations, and visibility constraints to replicate underwater conditions accurately.
\end{itemize}

By combining advanced simulation tools, a custom-designed aquaculture net pen model, and realistic sensor simulations, the environment provided a robust platform for evaluating AquaChat's capabilities under realistic conditions. Detailed experimental procedures and results will be discussed in a subsequent section.


\begin{figure}[t]
    \centering
    \includegraphics[width=1\linewidth]{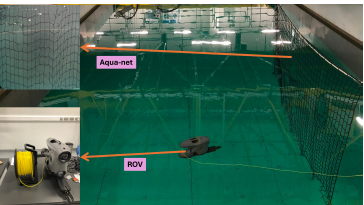}
    \caption{Real-time experiment setup for AquaChat: The configuration includes an ROV positioned to face the aqua-net, which is a rectangular net pen oriented vertically within a pool.}
    \label{fig:pool}
\end{figure}

Furthermore, we also show the experimental study on a real robotic platform. The experimental setup for AquaChat validation is depicted in the Figure~\ref{fig:pool}. It includes a pool containing a vertically installed net pen with dimensions of approximately 2.13 meters in width and 3.35 meters in height. The net pen serves as the primary structure for inspection and testing. The Blueye Pro ROV X developed by Blue Robotics \citep{blueye_x3}, is deployed in the pool, positioned to face the net pen. This configuration allows the ROV to navigate, inspect, and interact with the net pen, validating the AquaChat system's capabilities for real-time aquaculture net pen monitoring and analysis.

The Blueye Pro ROV X is a compact and robust underwater vehicle designed for versatile applications, including aquaculture monitoring and infrastructure inspection. Equipped with a 4K camera and powerful LED lights, it delivers high-quality visuals even in low-light conditions. The ROV can operate at depths of up to 305 meters (1,000 feet) and features high-performance thrusters for stability and maneuverability. Its modular design supports additional sensors and tools, while the user-friendly control system and Blueye Robotics SDK enable seamless operation, real-time monitoring, and custom application development \citep{blueye_sdk}.

\section{Results}\label{sec:res}

In this section, we present the experimental results of the AquaChat framework. First, we assess the performance of the LLM planner with various custom prompts and evaluate its responses. Then, we demonstrate the ROV's performance in executing the actions outlined in the generated plan during the inspection process.

\subsection{Performance of LLM methods}

To evaluate the performance of the LLM-based planner, a comparative analysis was conducted against a rule-based planner.
A rule-based planner operates by following predefined logical rules to perform specific tasks. It relies on a fixed set of conditions and actions, which are manually crafted for aquaculture inspection. Both planners were tested with diverse commands relevant to aquaculture net pen inspection tasks. The following metrics were considered for evaluation:

\begin{itemize}
    \item Plan Generation Time: The time taken to generate a complete plan.
    \item Task Understanding and Completeness: The planner’s ability to interpret and generate actionable plans for varied and complex commands.
    \item Flexibility: The range of instructions it could handle beyond predefined templates.

\end{itemize}

To evaluate the capabilities of the LLM-based planner for ROV operations, a diverse set of commands has been prepared to simulate real-world scenarios and test the system's ability to understand and execute tasks effectively. These commands include both structured instructions, which are clear and precise, and unstructured instructions, which mimic the variability and ambiguity of human communication. Structured commands define specific tasks with detailed parameters, such as movement patterns, inspection areas, and operational goals. On the other hand, unstructured commands challenge the LLM to infer intent and context, ensuring adaptability in less formal interactions. This approach ensures a comprehensive assessment of the planner's robustness and flexibility in translating user input into actionable tasks.

The following are examples of structured commands, where the tasks are clearly defined with specific actions and parameters:

\begin{itemize}
    \item     ``Inspect the entire net cage using a spiral method at a 3-meter distance.''
    \item  ``Move to the bottom-right corner of the net cage and capture an image.''
    \item  ``Detect net defects along the top edge of the cage.''
   \item   ``Perform a detailed inspection of the northern side of the net.''
    \item  ``Inspect the net cage from top to bottom and capture images at every meter.''
\end{itemize}

The following are examples of unstructured commands, which are less specific and may require interpretation or clarification:

\begin{itemize}
    \item     ``Can you check for holes in the net?''
     \item ``Go to the lower part and take pictures.''
     \item ``Scan the whole cage with high detail and tell me about defects.''
    \item  ``Take a close look at the east side and see if there are any damages.''
     \item ``Go around the net and find any issues.''
\end{itemize}

The experimental results showed in Table~\ref{tab:planner_comparison} highlight that the rule-based planner excelled in speed, generating plans within milliseconds for simple, predefined commands. However, its major limitations included such as command coverage. The planner could only handle a narrow set of commands. Complex or nuanced requests (e.g., "inspect along the top edge" or "scan the whole cage with high detail") resulted in unrecognized commands or inadequate plans. Adaptability is also another concern to rule based planner. The lack of contextual understanding prevented the planner from generating detailed or multi-step plans for ambiguous or sophisticated tasks.

On the other side, the LLM-based planner demonstrated significant advantages in terms of adaptability and task comprehension. It showed improvement in terms of command understanding. The LLM successfully generated detailed, multi-step plans for complex tasks. For example, commands like "Inspect the net cage from top to bottom and capture images at every meter" resulted in comprehensive plans involving movement, inspection, and defect detection. In addition, the planner could adapt to diverse phrasing and generate meaningful outputs for instructions that the rule-based planner failed to process. Plans included additional operations like defect detection and high-detail inspections, ensuring thorough task execution. However, the LLM-based planner exhibited certain limitations. The generation time ranged from 2.7 to 47.7 seconds, significantly slower than the rule-based approach. This could be attributed to the computational overhead of interpreting complex language instructions. For straightforward commands, such as ``Move to the bottom-right corner of the net cage and capture an image,'' the LLM produced more detailed plans than necessary, increasing time and complexity.

\begin{table}[h!]
\caption{Comparative analysis of rule-based and LLM-based planners.}
\label{tab:planner_comparison}
\begin{tabular}{p{2cm} p{2.5cm} p{2.5cm}}
\toprule

\textbf{Metric}              & \textbf{Rule-Based Planner} & \textbf{LLM-Based Planner} \\ \midrule
\textbf{Plan Generation Time} & \textasciitilde0.00–0.001 s & 2.7–47.7 s \\ \midrule
\textbf{Task Understanding}  & Limited                    & High                       \\ \midrule
\textbf{Command Flexibility} & Low                        & High                       \\ \midrule
\textbf{Task Completeness}   & Basic                      & Comprehensive              \\ \bottomrule
\end{tabular}

\end{table}

The results in Table~\ref{tab:comparison1} highlight the trade-offs between speed and versatility. While the rule-based planner is ideal for rapid execution of routine tasks, the LLM-based planner excels in handling complex, non-standard instructions that require contextual interpretation. For real-world aquaculture applications, the LLM-based planner's flexibility and depth make it a valuable tool, despite its slower response time. Optimizing the LLM for reduced latency could further enhance its applicability.

Table \ref{tab:comparison1} presents a comparative evaluation of command interpretation and plan generation by two different systems: LLM-based and Rule-based planners. It categorizes user commands into structured and unstructured types and lists their corresponding expected plans, outputs generated by each system, and their correctness. For example, the structured command ``Inspect the western side'' is expected to produce the plan inspect(west). While the LLM system correctly generates the expected output (indicated by a checkmark), the Rule-based system fails. Similarly, for unstructured commands like ``Take a close look at the west side'' and ``Can you scan the whole net?'', only the LLM system produces the correct output, demonstrating its superior ability to handle varied and less formal language. This result effectively highlights the differences in performance and accuracy between the two systems for structured and unstructured command processing.

\begin{table*}[ht]
    \centering
    \caption{Comparison of Command Outputs for LLM and Rule-Based Planners. }
    \label{tab:comparison1}
    \begin{tabular}{p{2cm} p{2cm} p{2.8cm} p{2cm} p{2cm} p{2cm}}
        \toprule
        \textbf{Type} & \textbf{User Command} & \textbf{Expected Plan} & \textbf{Output (LLM)} & \textbf{Output (Rule-Based)} & \textbf{Correct (LLM/Rule)} \\ \midrule
        Structured            & "Inspect the western side." & \texttt{inspect\_net(west)} & \ding{51} & \ding{55} & LLM \ding{51} / Rule \ding{55}  \\
        Unstructured          & "Take a close look at the west side." & \texttt{inspect\_net(west)} & \ding{51} & \ding{55}  & LLM \ding{51} / Rule \ding{55} \\
        Unstructured          & "Can you scan the whole net?" & \texttt{inspect\_net()} & \ding{51} & \ding{55} & LLM \ding{51} / Rule \ding{55} \\
        \bottomrule
    \end{tabular}
    
\end{table*}

\begin{figure}[h]
    \centering
    \includegraphics[width=1\linewidth]{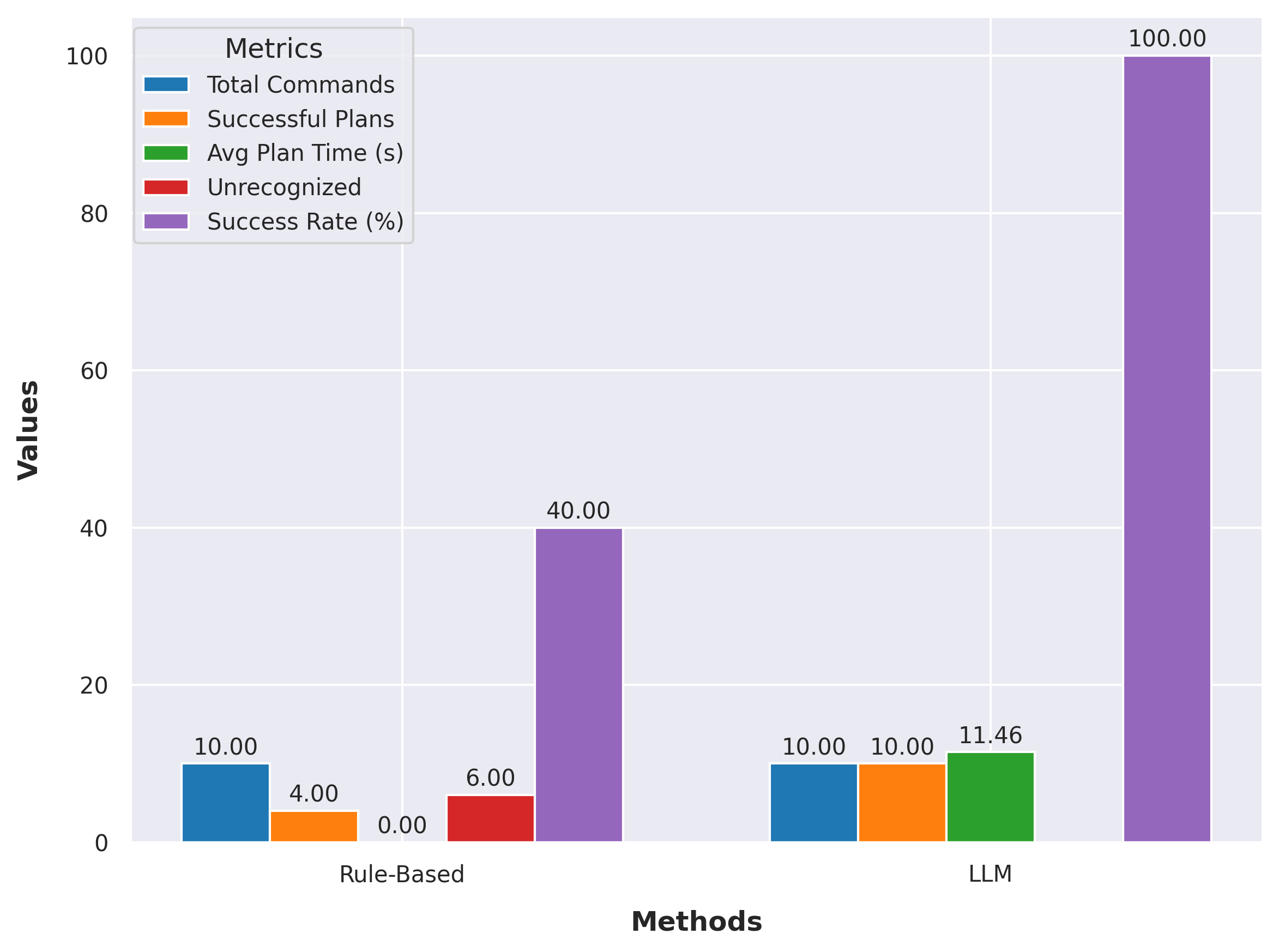}
    \caption{Comparison of Metrics Across Methods: The bar chart illustrates the performance differences between Rule-Based and LLM-based methods across five key metrics: total commands, successful plans, average plan generation time, unrecognized commands, and success rates.   }
    \label{fig:qua-res-planners}
\end{figure}

Table~\ref{tab:comparison2} presents the performance results for each prompt in terms of plan success rate, execution rate, and call time. The rule-based planner performs well with structured prompts but fails to generate valid plans for more complex or unstructured commands. In contrast, the LLM-based planner demonstrates a higher success rate and successfully generates the desired missions for all prompts. Although it requires more time for plan generation, the LLM planner excels in handling complexity and processing human natural language. Similarly, the average results for both planners are plotted in Figure~\ref{fig:qua-res-planners}, where key metrics such as total commands, successful plans, average plan generation time, and success rates are compared. This visualization clearly highlights the differences in performance between the Rule-Based and LLM-Based planners, with numerical values annotated on top of each bar for easy reference. The plot provides a detailed overview of how each planner performs across these metrics, offering valuable insights into their respective efficiencies and effectiveness for aquaculture net pen inspection tasks.

\begin{table*}[ht]
\centering
\caption{Comparison of Rules-based Plan and LLM-based Plan. PSR: planning success rate, EXESR: execution success rate, Time: LLM time (s)}
\label{tab:comparison2}

\begin{tabular}{p{6cm}cccccc}
\toprule
\multirow{2}{*}{\textbf{Prompt}} & \multicolumn{3}{c}{\textbf{Rules-based Plan}} & \multicolumn{3}{c}{\textbf{LLM-based Plan}} \\ \cline{2-7} 
                                 & \textbf{PSR \%} & \textbf{EXESR \%} & \textbf{\# Time (s)} & \textbf{PSR \%} & \textbf{EXESR \%} & \textbf{\# Time (s)} \\ \midrule

\scriptsize Inspect the entire net cage using a spiral method at a 3-meter distance & 100 & 100 & 0& 100 & 100 & 3.2 \\ \midrule
\scriptsize Move to the bottom-right corner of the net cage and capture an image & 100 & 90 & 0& 100 & 95 & 4.1 \\ \midrule
\scriptsize Detect net defects along the top edge of the cage & 100 & 85 & 0& 100 & 90 & 3.8 \\ \midrule
\scriptsize Perform a detailed inspection of the northern side of the net & 100 & 80 & 0& 100 & 95 & 3.4 \\ \midrule
\scriptsize Inspect the net cage from top to bottom and capture images at every meter & 90 & 85 & 0 & 100 & 90 & 3.1 \\\midrule

\scriptsize Can you check for holes in the net? & 0 & 0 & 0 & 65 & 95 & 4.4 \\ \midrule
\scriptsize Go to the lower part and take pictures& 0 & 0 & 0 & 65 & 95 & 4.5 \\ \midrule
\scriptsize Scan the whole cage with high detail and tell me about defects& 0 & 0 & 0 & 90 & 60 & 5.4 \\ \midrule
\scriptsize Take a close look at the east side and see if there are any damages& 0 & 0 & 0 & 70 & 95 & 3.1 \\ \midrule
\scriptsize Go around the net and find any issues& 0 & 0 & 0 & 65 & 95 & 4.4 \\ 

\bottomrule

\end{tabular}
\end{table*}


\subsection{Performance of Control method}
In this section, we evaluate the performance of the control method employed in the AquaChat framework, focusing on key performance benchmarks such as navigation accuracy and efficiency. The analysis examines the system's ability to follow planned trajectories, maintain stability in dynamic underwater environments, and execute inspection tasks effectively. Additionally, metrics such as positional error is used to assess the overall effectiveness of the control method in achieving the desired mission objectives.

\begin{figure*}[t]
    \centering
    \includegraphics[width=1\linewidth]{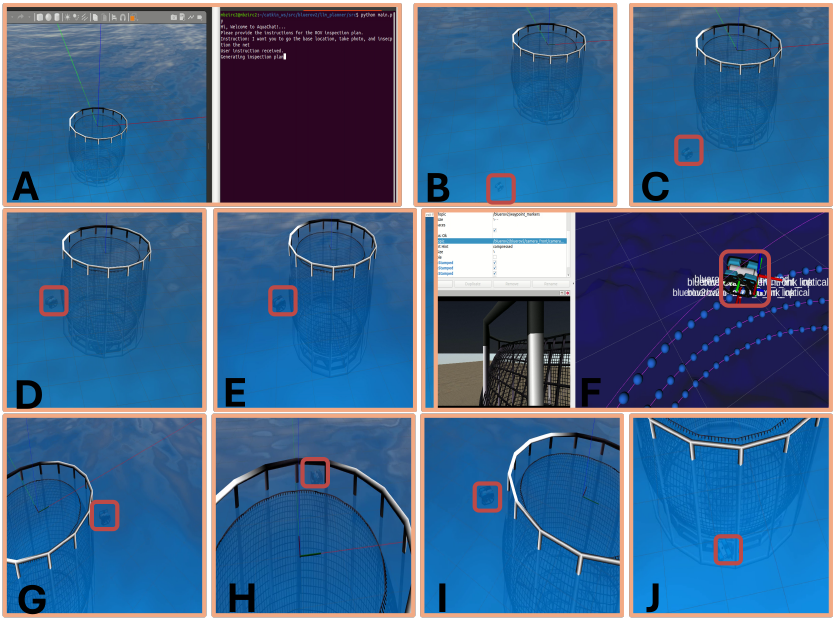}
    \caption{Showcasing plan executing in a simulation for actions in the plan e.g ``move to'', ``inspect''.}
    \label{fig:vis-res}
\end{figure*}

The results are presented for two distinct scenarios: ``Move To'' and ``Inspect''. In the ``Move To'' scenario, the ROV is asked to navigate towards a specific location using precise trajectory-following algorithms. Metric such as path accuracy with respect to the reference path is analyzed to evaluate performance. In the ``Inspect'' scenario, the focus shifts to the ROV's ability to perform detailed inspections, such as capturing images or moving around the net. A detailed visualization of these action execution is shown in Figure~\ref{fig:vis-res}.

\begin{figure}[h]
    \centering
    \includegraphics[width=1\linewidth]{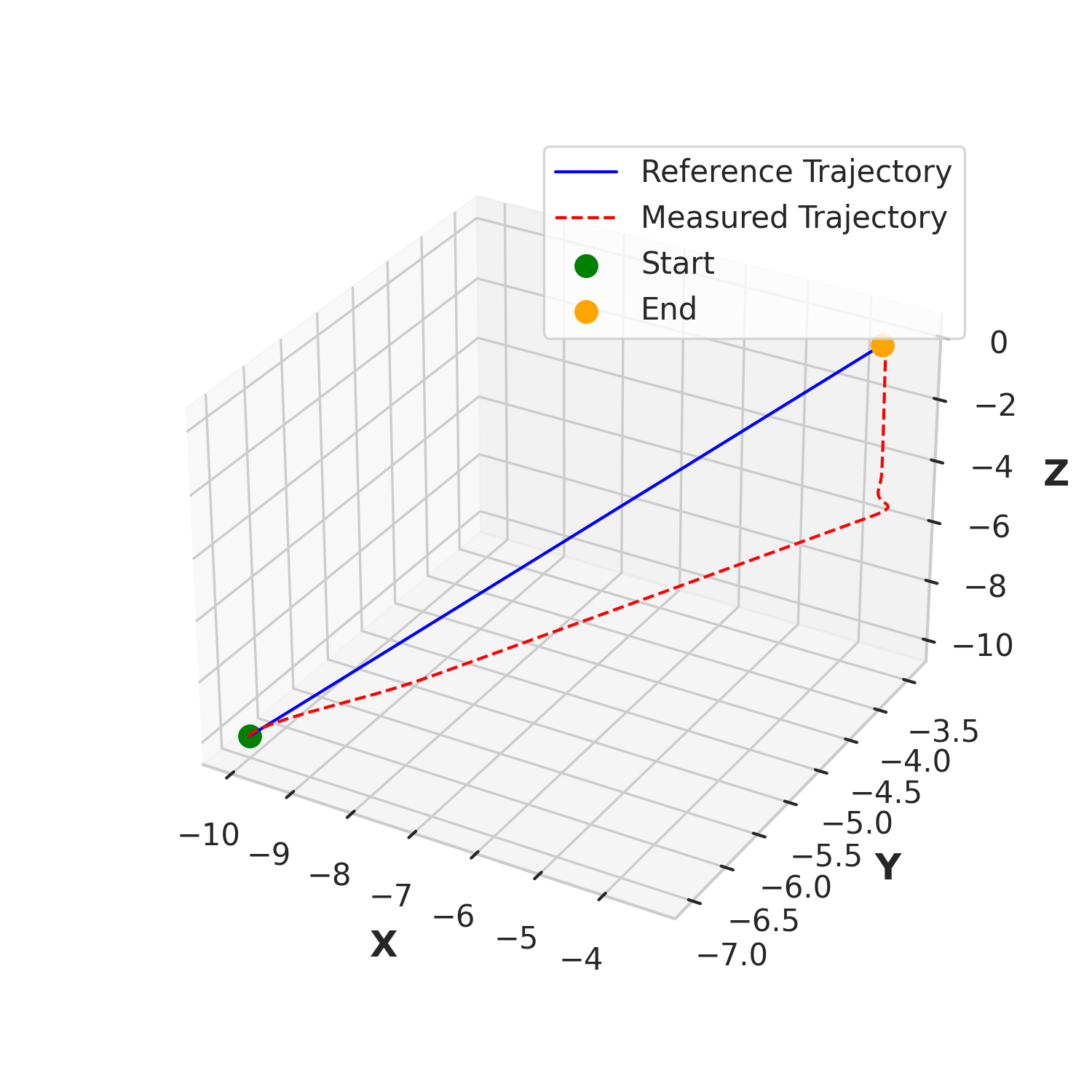}
    \caption{Case 1 (``move to'' ): ROV 3D position during the plan execution.}
    \label{fig:move1}
\end{figure}

Figure~\ref{fig:move1} illustrates a 3D trajectory comparison between a reference and a measured path over positions $x,y$ and $z$. The reference trajectory is depicted by a solid blue line, while the measured trajectory is shown with a dashed red line. Both paths originate from the same starting point, marked by a green dot at approximately $(x: -10, y:-7, z:-10)$. The endpoints are indicated by an orange dot, showing the paths converge at a similar destination in 3D space. Though the measured trajectory follows the general direction and pattern of the reference, it deviates slightly, especially along the path, before aligning closely with the final destination $(x: -3.5, y:-3.5, z:0)$. Overall, the results demonstrates the effectiveness of the tracking process and the ability to perform the aqua-net inspection tasks.

\begin{figure}[h]
    \centering
    \includegraphics[width=1\linewidth]{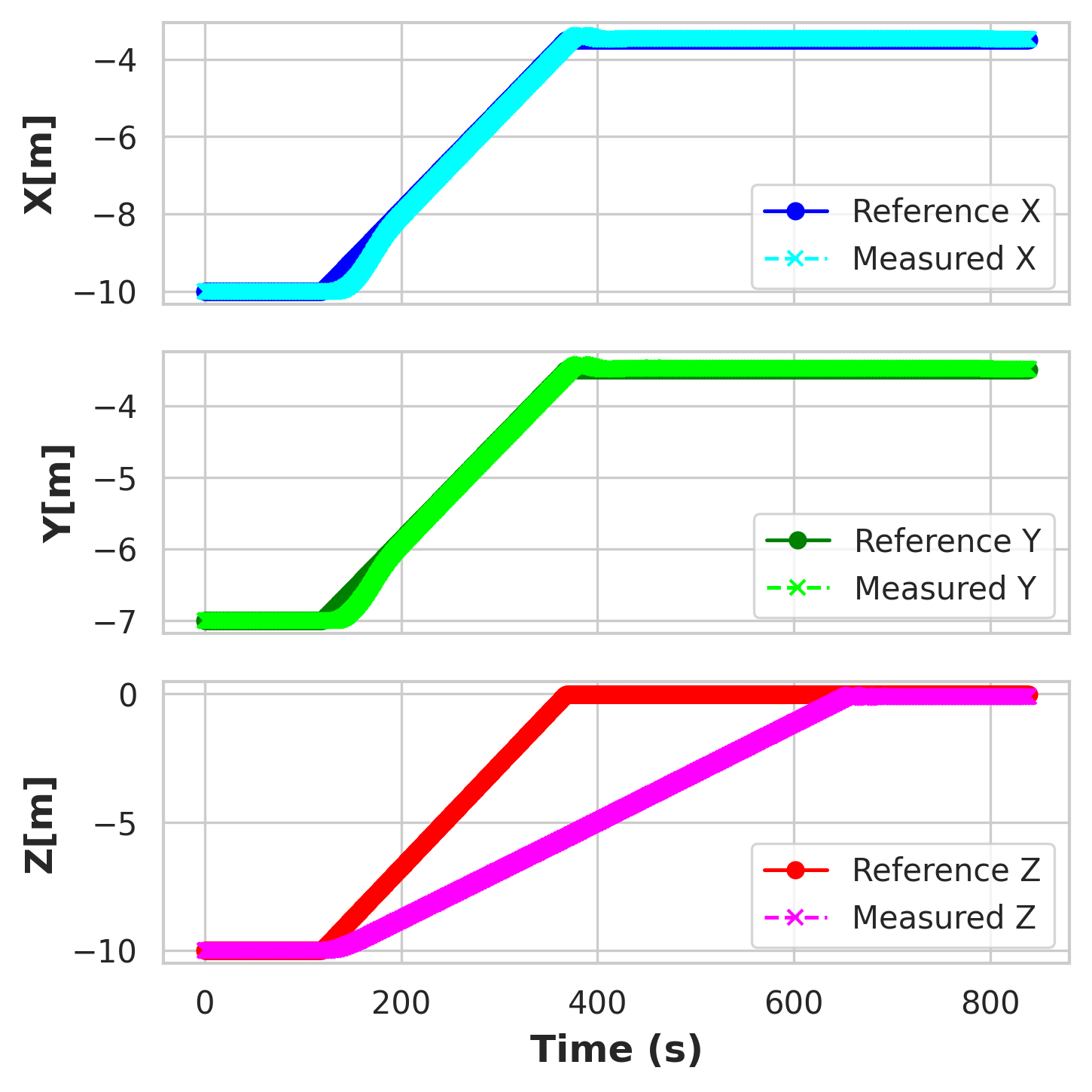}
    \caption{Case 1 (``move to'' ): ROV position (x, y, z) during the plan execution.}
    \label{fig:move2}
\end{figure}

Figure~\ref{fig:move2} presents the reference and measured positions over time for three spatial coordinates: X, Y, and Z. Each coordinate is shown in its own subplot. In the top plot, the X coordinate's reference (shown in blue) starts at -10m and increases to about -4m before stabilizing. The measured X position (depicted in cyan) follows this trajectory closely, indicating effective tracking with only minor deviations. In the middle subplot, the Y coordinate's reference (in green) begins at -7 meters and rises to nearly -3.5m, where it remains constant. The measured Y position (in light green) matches the reference well, suggesting successful tracking with minimal errors. Finally, the bottom subplot illustrates the Z coordinate, with the reference trajectory (in red) moving from -10m to around 0m before leveling off. The measured Z position (shown in magenta) closely parallels the reference, showing slight discrepancies. Overall, the measured data mirrors the reference data well across all coordinates, reflecting successful tracking performance over the 850-second time period.

\begin{figure}[h]
    \centering
    \includegraphics[width=1\linewidth]{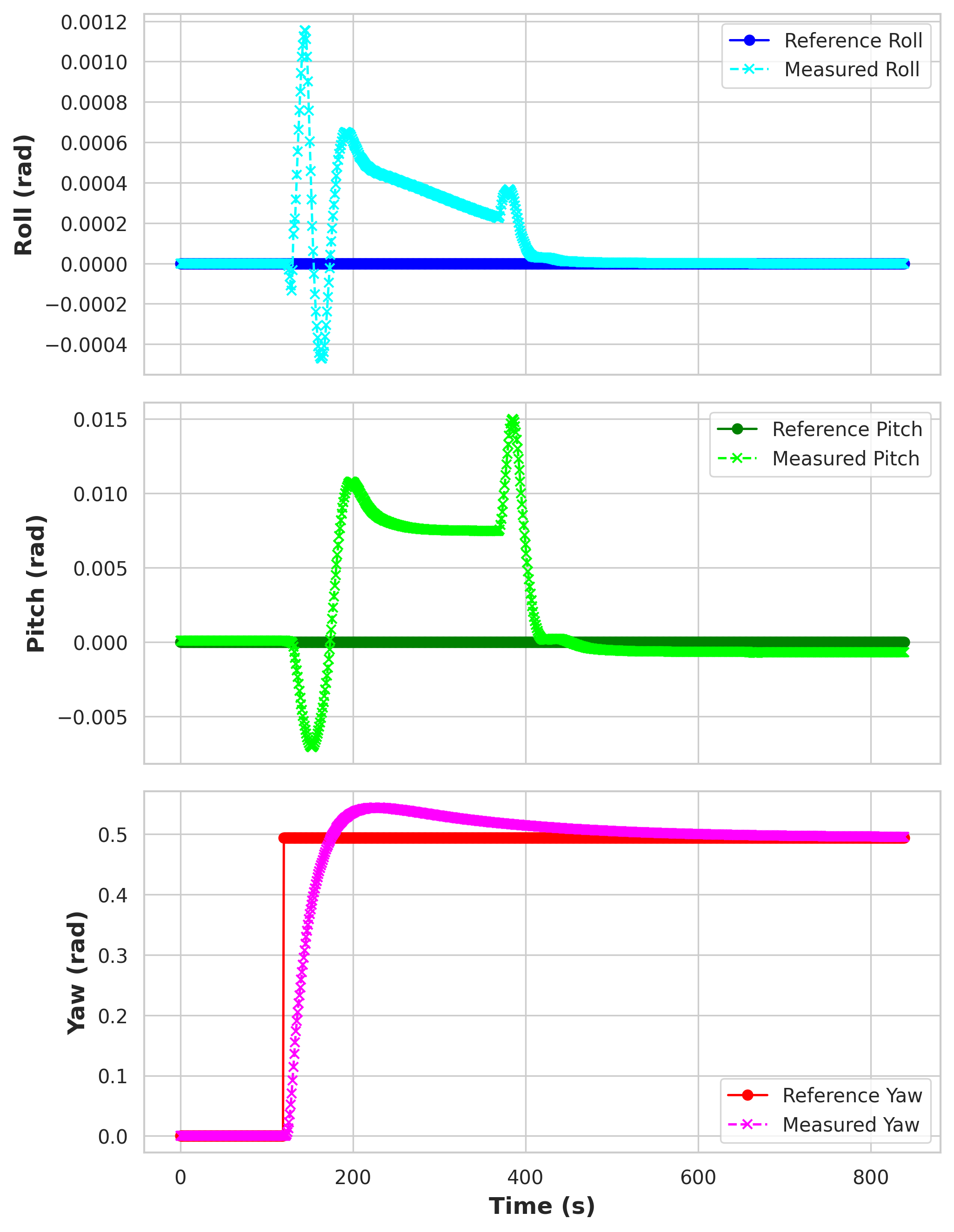}
    \caption{Case 1 (``move to'' ): ROV orientation (roll, pitch, yaw) during the plan execution.}
    \label{fig:move3}
\end{figure}

Figure~\ref{fig:move3} displays the reference and measured angular orientations over time for roll, pitch, and yaw, each in separate subplots. For roll, the reference is a stable blue line near zero, while the measured roll (cyan) shows initial oscillations before stabilizing. The pitch plot shows a steady reference (green) around zero, with the measured pitch (light green) fluctuating initially before aligning closely with the reference. The yaw graph illustrates a reference (red) that increases sharply to 0.5 radians and remains constant, while the measured yaw (magenta) initially lags but eventually aligns closely. Overall, the measured data follows the reference with initial discrepancies that stabilize over time.

\begin{figure}[h]
    \centering
    \includegraphics[width=1\linewidth]{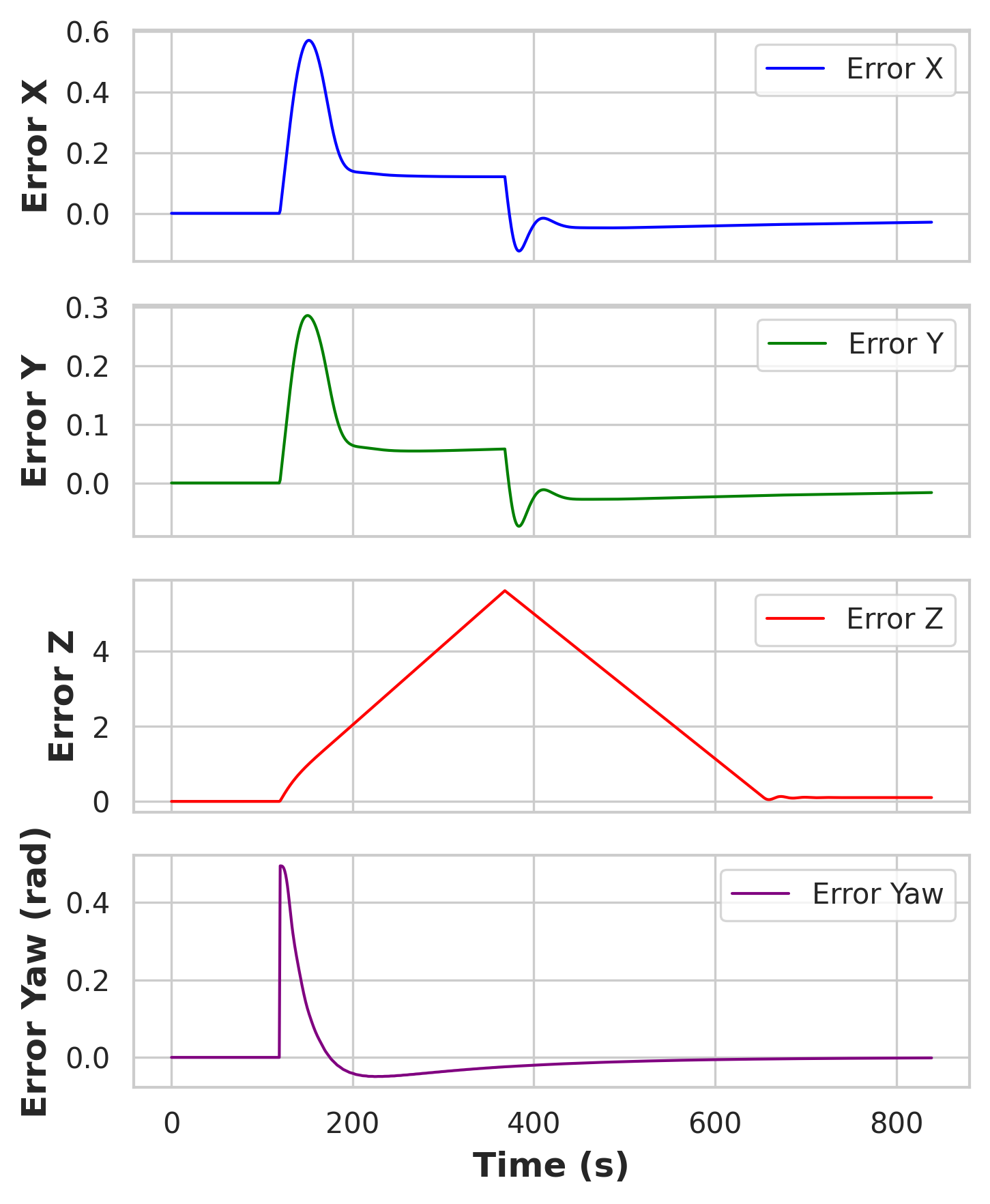}
    \caption{Case 1 (``move to'' ): ROV position error (x, y, z, yaw) during the plan execution.}
    \label{fig:move4}
\end{figure}

\begin{figure}[h]
    \centering
    \includegraphics[width=1\linewidth]{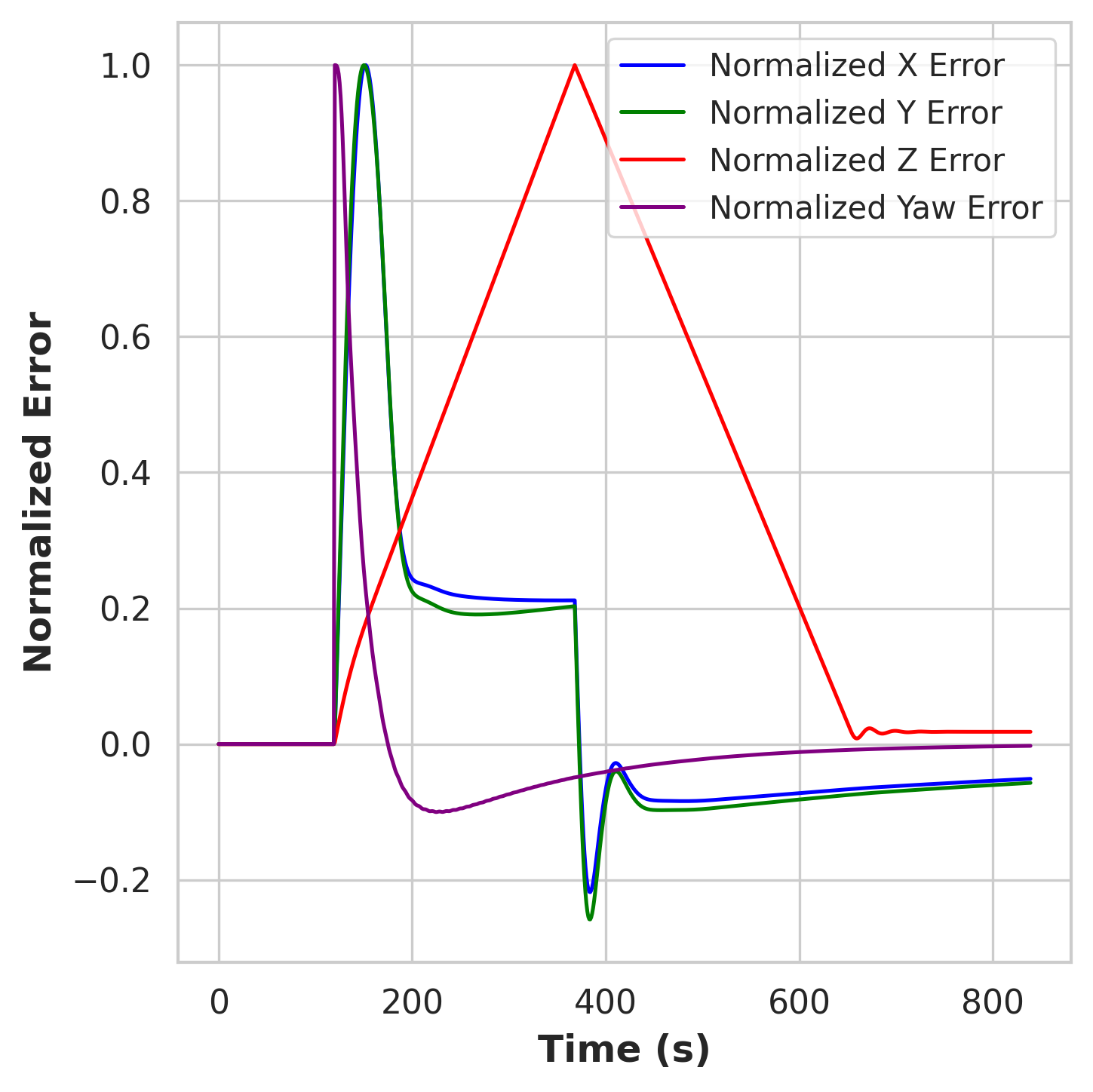}
    \caption{Case 1 (``move to'' ): ROV normalized error (x, y, z, yaw) during the plan execution.}
    \label{fig:mov5}
\end{figure}

Figure~\ref{fig:move4} depict the errors in X, Y, Z positions, and yaw over time, alongside normalized errors in Figure~\ref{fig:mov5}. Initially, each component shows significant peaks—most notably Z, which exceeds 4 units. Over time, these errors decrease and stabilize, with X and Y errors peaking around 0.5 and 0.3, respectively, and yaw error quickly diminishing from a peak near 0.5 radians. The normalized error plot highlights the relative magnitudes and demonstrates that, despite initial spikes, errors converge towards minimal values, indicating improved accuracy over the observed period.

\begin{figure}[h]
    \centering
    \includegraphics[width=1\linewidth]{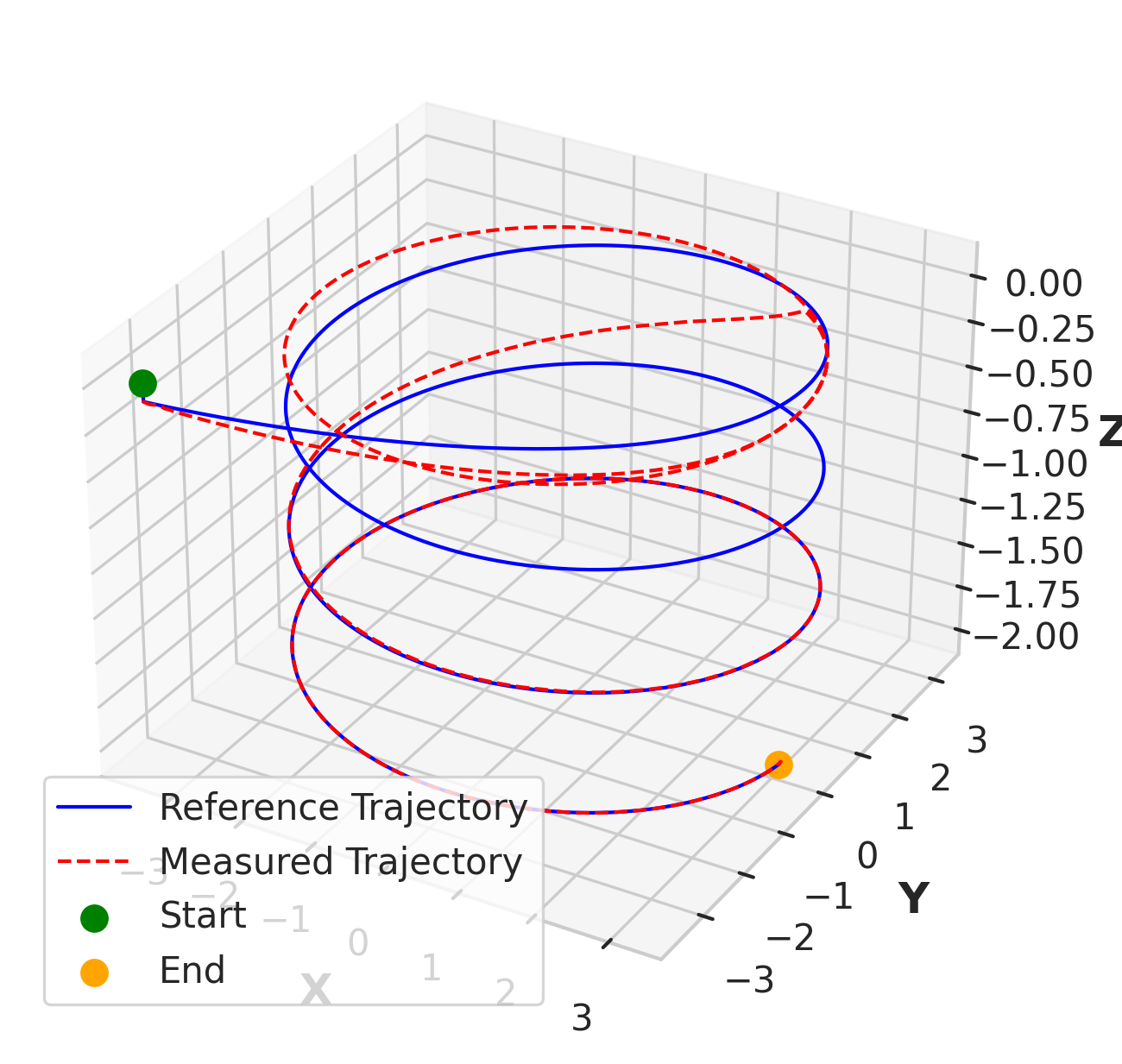}
    \caption{Case 2 (``inspect''): ROV 3D position during the plan execution. }
    \label{fig:inspect1}
\end{figure}

Figure~\ref{fig:inspect1} illustrates a 3D spiral trajectory with both reference and measured paths. The reference trajectory is shown as a smooth blue curve forming a helix with a radius of $3.5m$, and heading towards the origin, starting from a green marker and ending at an orange marker. The measured trajectory, depicted by a dashed red line, closely follows the reference path, though with slight deviations. The trajectory spirals downwards along the Z-axis while simultaneously looping in the X-Y plane. Starting at the top of the helix and ending at the bottom, the paths indicate that the measured trajectory effectively tracks the reference, maintaining alignment with minor discrepancies throughout the descent. This visualization underscores accurate tracking performance in a complex spatial path.

\begin{figure}[h]
    \centering
    \includegraphics[width=1\linewidth]{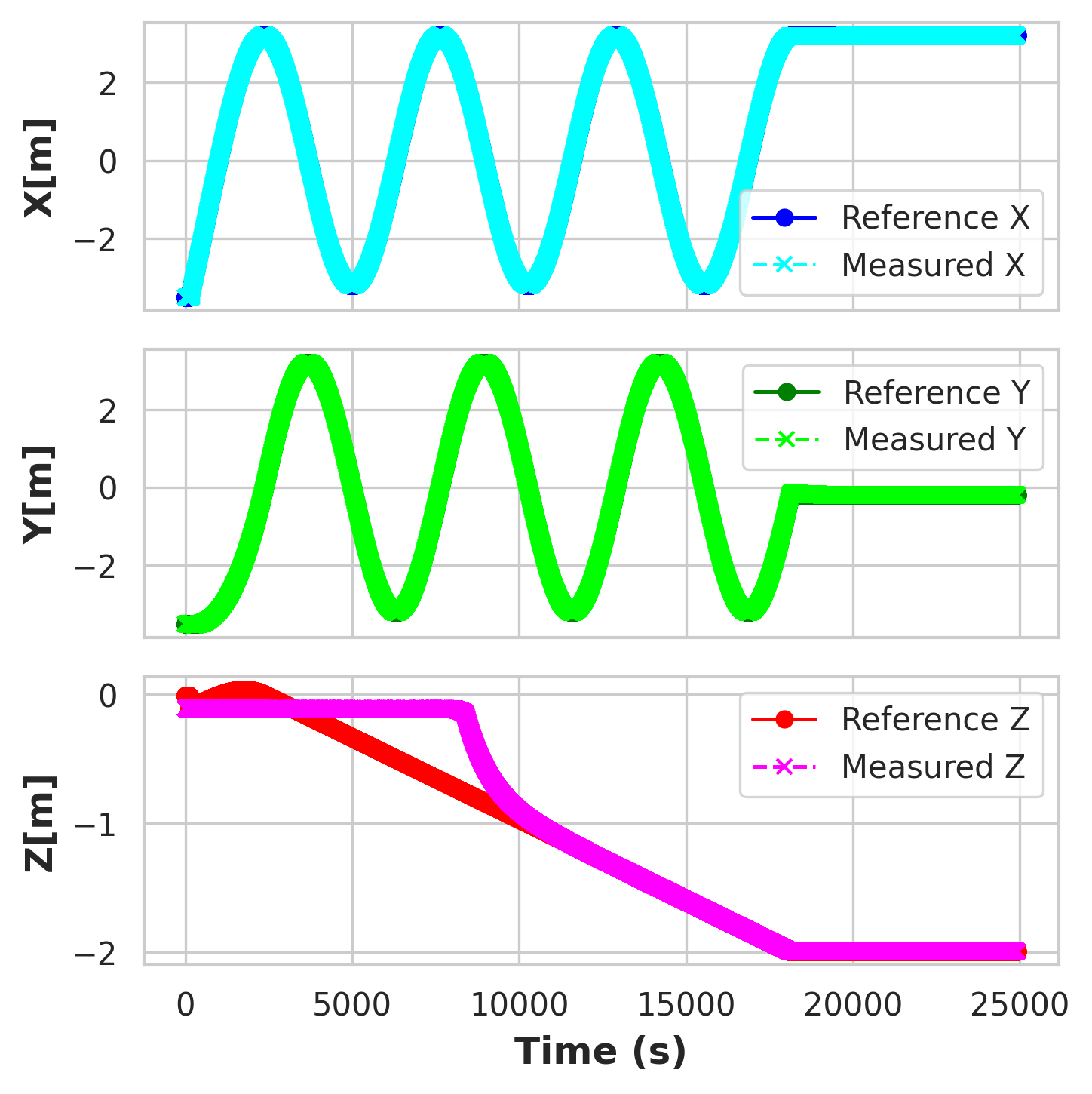}
    \caption{Case 2 (``inspect'' ): ROV position (x, y, z) during the plan execution.}
    \label{fig:inspect2}
\end{figure}

Figure~\ref{fig:inspect2} depict the reference and measured positions over time for $x, y$ and $z$ coordinates. Both the $x$ and $y$ plots show sinusoidal patterns with measured trajectories (cyan and light green dashed lines) closely following their respective references (solid blue and green lines). The oscillations occur within a range of approximately $<-2,2m>$, demonstrating accurate tracking of the oscillatory path. In contrast, the $z$ plot presents a downward linear trajectory. The measured $z$ (magenta dashed line) initially aligns closely with the reference (solid red line) but shows slight deviations during the descent before stabilizing near the end. Overall, the plots illustrate effective tracking performance for the oscillating $x$ and $y$ trajectories and a gradual alignment in the $z$ trajectory.

\begin{figure}[h]
    \centering
    \includegraphics[width=1\linewidth]{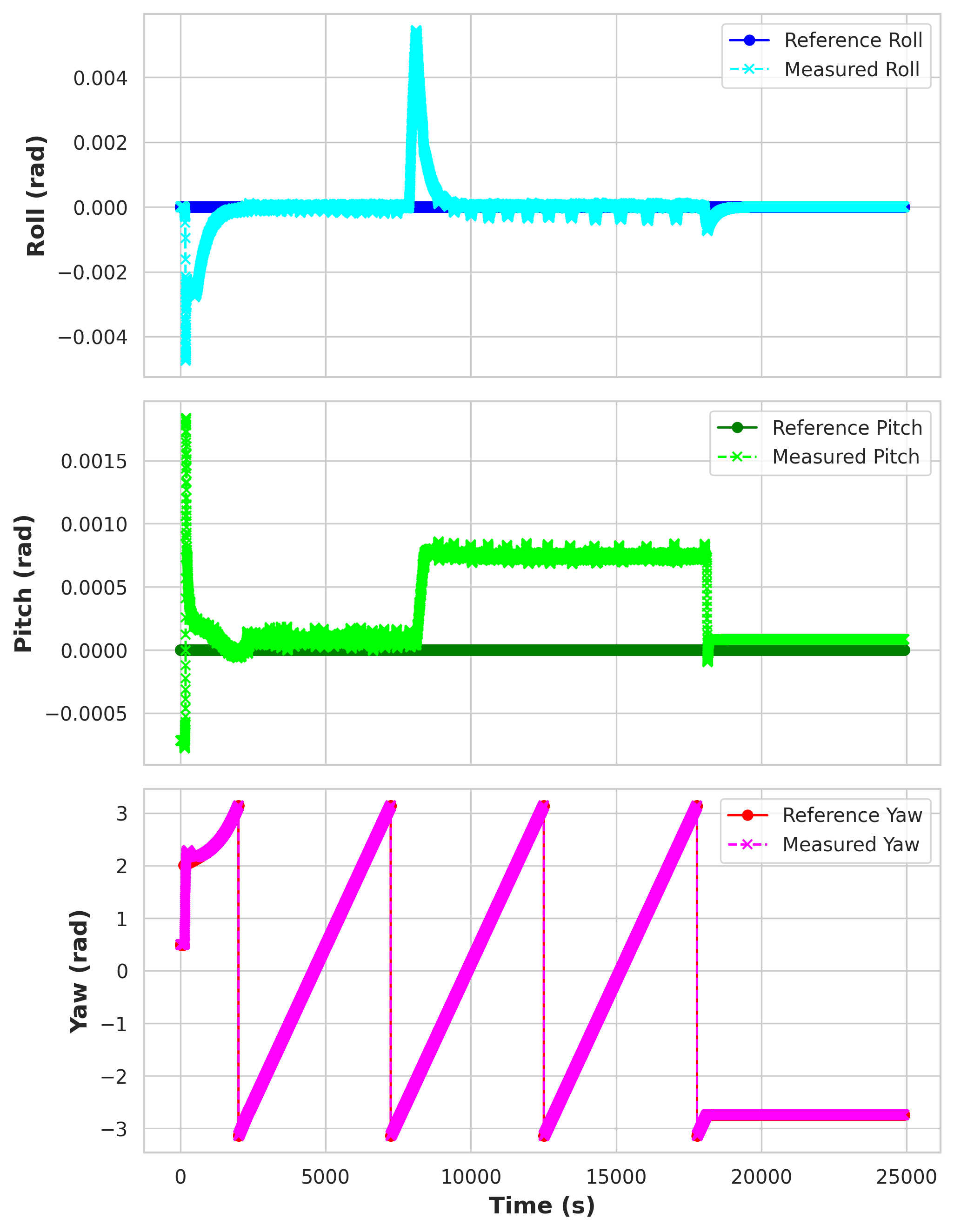}
    \caption{Case 2 (``inspect''): ROV orientation (roll, pitch, yaw) during the plan execution.}
    \label{fig:inspect3}
\end{figure}

Figure~\ref{fig:inspect3} show the reference and measured angular orientations over time for roll, pitch, and yaw. The roll plot has a stable reference (blue line) near zero, while the measured roll (cyan) includes a noticeable spike before stabilizing with minor oscillations. The pitch plot shows a stable reference (green line) at zero, with the measured pitch (light green) initially diverging before aligning steadily with the reference. The yaw plot features a reference (red line) exhibiting periodic angular changes between $<-3,3>$ radians, tracked closely by the measured yaw (magenta dashed line) with sharp, consistent transitions. Overall, the plots reflect effective tracking with some initial deviations and consistency in cyclical yaw angle behavior.

\begin{figure}[h]
    \centering
    \includegraphics[width=1\linewidth]{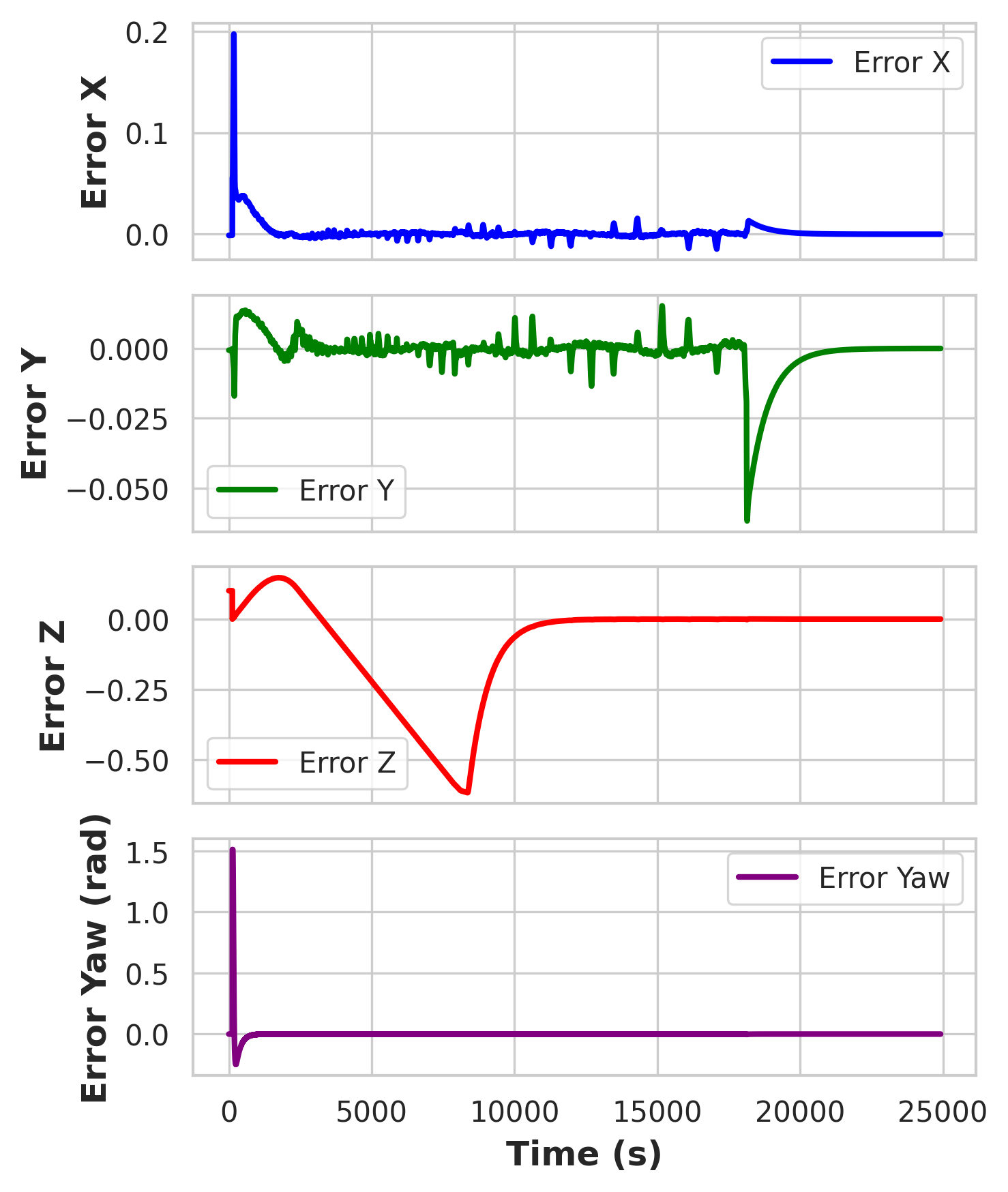}
    \caption{Case 2 (``inspect'' ): ROV position error (x, y, z, yaw) during the plan execution.}
    \label{fig:inspect4}
\end{figure}

\begin{figure}[h]
    \centering
    \includegraphics[width=1\linewidth]{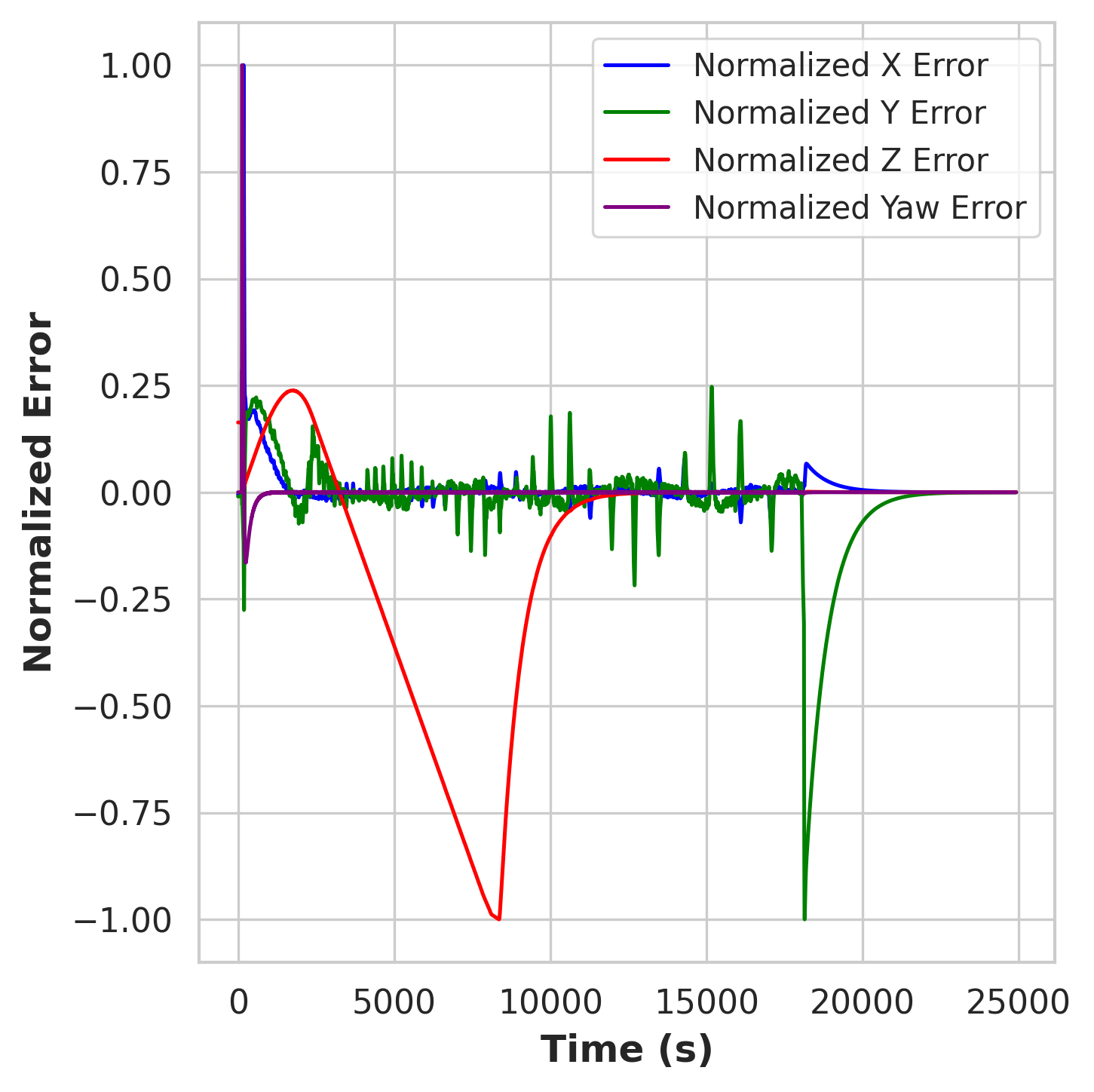}
    \caption{Case 2 (``inspect'' ): ROV normalized error (x, y, z, yaw) during the plan execution.}
    \label{fig:inspect5}
\end{figure}

Moreover, the errors during the tracking are shown in Figures~\ref{fig:inspect4} and~\ref{fig:inspect5}. The first set of plots shows the errors in $x, y, z$ positions, and yaw over time. The $x$ error (blue) starts around 0.2 and stabilizes near zero. The $y$ error (green) fluctuates slightly around zero throughout. The $z$ error (red) dips significantly before stabilizing, while the yaw error (purple) quickly falls to near zero. The second plot illustrates normalized errors for each component, highlighting their relative magnitudes. Initially, yaw shows the largest error, which reduces rapidly. Positions $x$ and $y$ exhibit notable fluctuations, but all errors converge towards zero over time, indicating improved tracking accuracy.

The obtained results provides insight into the spatial and angular tracking performance of the proposed system during the inspection mission. The analysis shows that while initial deviations exist, particularly in yaw and $x$ position, the system demonstrates the capability to stabilize and closely follow reference trajectories over time.

For the $x$ and $y$ coordinates, the system exhibits effective oscillatory tracking, with initial errors that decrease and maintain minimal levels. The consistency in sinusoidal patterns emphasizes robust control performance. The $z$ position initially presents more pronounced errors, as shown by significant deviations. However, stabilization occurs, and errors decrease, indicating that the system adapts and corrects over time.

The results from the ROV orientation during the inspection reveal that roll and pitch maintain low error levels after initial fluctuations. The yaw angle, initially showing substantial error, quickly aligns with the reference trajectory, illustrating rapid correction capabilities. The cyclic behavior in yaw tracking suggests the system can manage repetitive angular movements effectively.

The normalized error plots highlight the relative magnitudes of errors across different dimensions. Despite initial spikes, especially in yaw and $z$, the normalized errors converge towards zero. This convergence suggests successful calibration and adaptation mechanisms are in place, reducing discrepancies over time.

The results suggest promising tracking accuracy for applications requiring precise spatial and angular control in aquaculture environment. However, initial discrepancies, particularly in $z$ positions and yaw, highlight areas for further refinement. Future work may focus on enhancing initial calibration and predictive correction algorithms to improve early-phase accuracy. Enhancements in these aspects could lead to even more robust tracking performance for dynamic and complex paths.

\begin{figure}[h]
    \centering
    \includegraphics[width=1\linewidth]{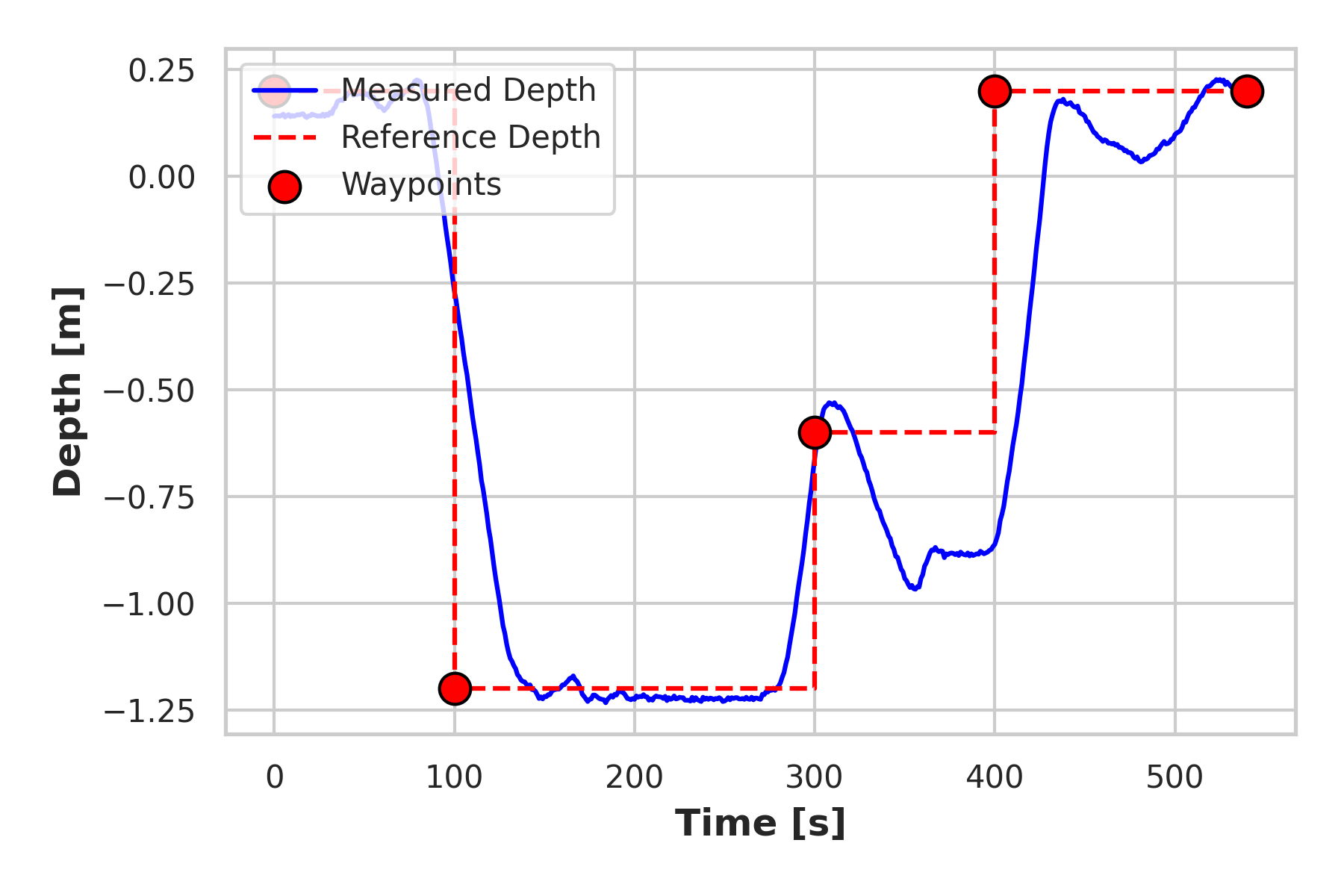}
    \caption{Experiment 1: ROV depth profile over time.}
    \label{fig:depth1}
\end{figure}

\begin{figure}[h]
    \centering
    \includegraphics[width=1\linewidth]{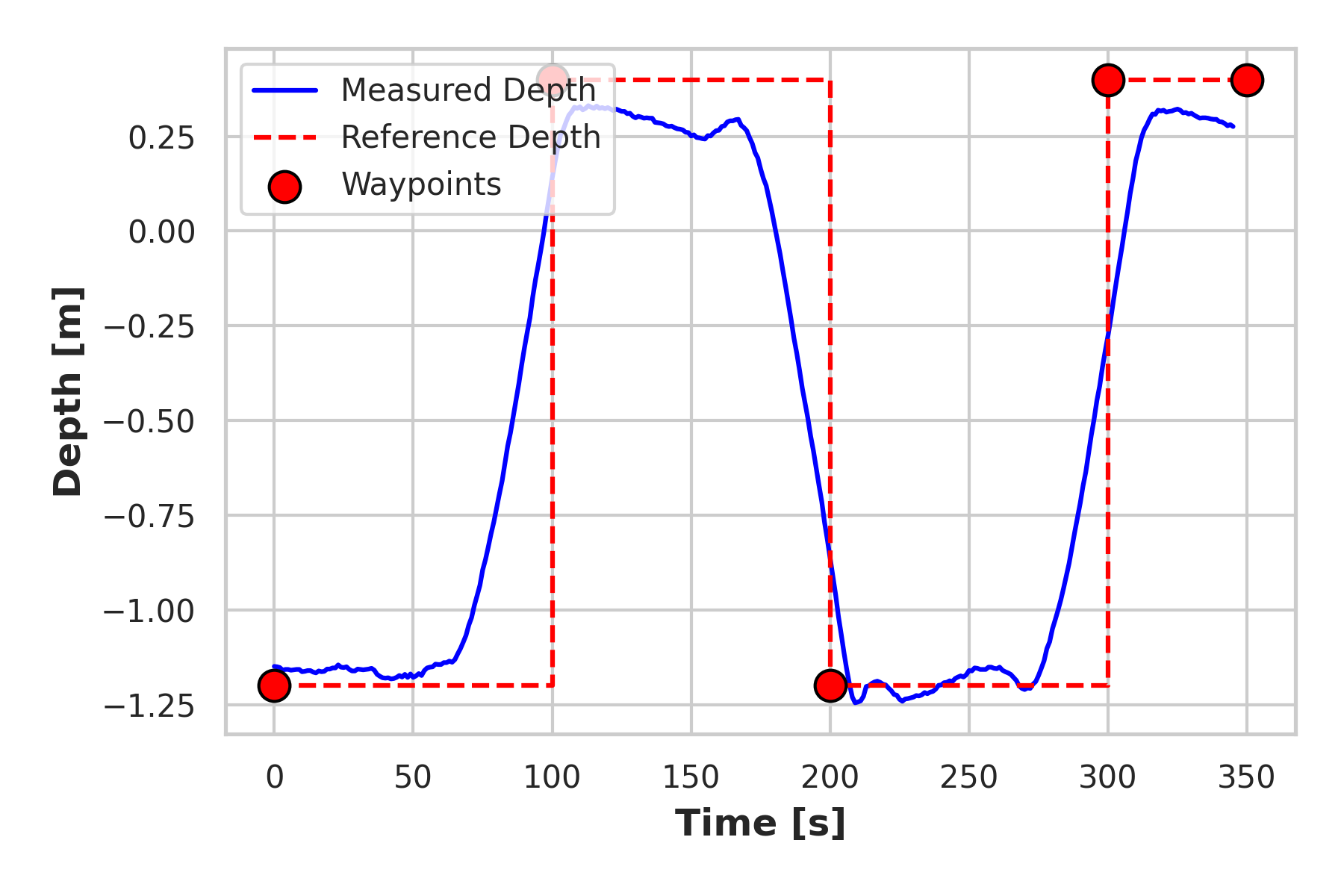}
    \caption{Experiment 2: ROV depth profile over time.}
    \label{fig:depth2}
\end{figure}

To further validate the AquaChat system, we conducted experiments using the Blueye Pro ROV X to inspect a vertically installed aquaculture net pen. The ROV was asked to follow a predefined zig-zag inspection trajectory, defined by a sequence of waypoints that guided its movement across the net surface. This pattern ensures comprehensive coverage of the inspection area while maintaining efficiency.

During the experiment, the ROV's forward, lateral, and downward movements were controlled by applying absolute force values within a normalized range of $(0-1)$. This allowed precise adjustment of the ROV's thrust for stable and accurate navigation, even in challenging conditions. Depth profiles were recorded during the operation, demonstrating the ROV's ability to maintain consistent depth and follow the planned trajectory accurately. The data obtained highlights the system's potential for efficient net pen monitoring, with results visualized through depth profile plots that align closely with the planned inspection pattern.

The experimental results are illustrated through two experiments. Figure~\ref{fig:depth1} shows experiment 1 that compares the measured depth data (blue line) with the reference zig-zag pattern (red dashed line and waypoints marked by bold red circles). The reference pattern follows a step-like trajectory, beginning at -1.25m until 100 seconds, sharply rising to 0.25m between 100 and 200 seconds, descending back to -1.25m until 300 seconds, and finally ascending again to 0.25 meters until 500 seconds. The measured depth exhibits fluctuations within a similar range, demonstrating the ROV's capability to approximate the reference path.

Figure~\ref{fig:depth2} provides a time-series comparison of the same data for experiment 2, highlighting deviations from the reference trajectory. The measured depth closely tracks the zig-zag pattern, with fluctuations evident between -1.25m and 0.25m. The visualization emphasizes the alignment and occasional deviations of the ROV's motion relative to the predefined waypoints, aiding in the evaluation of control performance. The obtained results demonstrate the applicability of the AquaChat for the real underwater platform in the aquaculture net pens inspection. 

\section{Discussion and Future Directions}

In this section, we explore the implications of the proposed framework, identify challenges faced during the implementation, and suggest potential aspects for future research and development.

The underwater environment presents inherent complexities, such as uneven lighting, dynamic currents, and varying levels of biological activity. These factors can introduce noise and uncertainty in perception and navigation systems. Future work should focus on developing robust sensing techniques that make use of advanced computer vision and sensor fusion methods to improve environmental understanding and mitigate these challenges.

Turbulence and poor visibility due to sedimentation or plankton blooms significantly affect the ROV's ability to inspect aquaculture net pens effectively. Addressing these issues may involve integrating acoustic sensors, sonar imaging, and AI-enhanced filtering methods to provide reliable data despite adverse conditions. Additionally, adaptive control algorithms could help the ROV maintain stability and precise movements in turbulent waters.

While the current framework utilizes general-purpose LLMs for high-level planning, the models may not perform optimally in niche domains without customization. Research into fine-tuning LLMs on domain-specific datasets related to underwater navigation and aquaculture could improve their contextual understanding and decision-making. Incorporating real-world inspection data and simulated scenarios into the training pipeline could further enhance adaptability and performance.

Customizing LLMs for specific underwater tasks involves tailoring their language understanding to accommodate technical and context-specific commands, and domain-relevant ontologies. Future efforts should explore techniques such as prompt engineering, model pretraining with domain-specific knowledge, and integrating LLMs with external knowledge bases to enhance their utility in complex underwater operations.

Scalability remains a critical consideration as aquaculture operations expand. The framework should be capable of handling larger net pen structures, requiring more extensive inspection routes and increased computational demands. Employing distributed planning algorithms and leveraging edge computing technologies could address scalability concerns. Further, designing modular ROV systems with interchangeable components might enable flexible scaling.

The current framework focuses on single-ROV operations within standard-sized net pens. Future work could investigate the coordination of multiple ROVs for simultaneous inspection tasks, reducing overall inspection time and increasing efficiency. This could involve developing multi-agent planning algorithms, real-time communication protocols, and strategies for conflict resolution among ROVs. Extending the approach to inspect larger and more complex net pen systems would require integrating advanced path-planning algorithms and hierarchical control architectures.

By addressing the outlined challenges and pursuing the proposed future directions will enable the development of a more robust, adaptable, and scalable framework for underwater inspection in aquaculture environments. These advancements could contribute significantly to the efficiency and sustainability of aquaculture operations.

\section{Conclusion}\label{sec:conc}

This work proposed a novel framework named AquaChat, using LLM-guided navigation coupled ROV for the inspection of aquaculture net pens. By integrating high-level planning with domain-specific tasks and real-time sensing, the system demonstrates the potential for enhancing underwater inspection efficiency and precision. The modular design of the framework enables adaptability to various operational scenarios, addressing challenges such as environmental complexity. Moreover, it offers a user-friendly and ease-of-use aquaculture net pen inspection via LLM planner by taking instruction in a natural language format. Overall, the proposed approach lays a strong foundation for advancing the role of AI-driven technologies in underwater inspection and beyond.
In future, the framework will be enhanced to address challenges including model adaptation, scalability, and the extension to multi-ROV systems. Addressing these challenges will unlock new opportunities for improving aquaculture management and sustainability.

\section*{Acknowledgement}
\noindent This work is supported by the Khalifa University under Award No. RC1-2018-KUCARS-8474000136, CIRA-2021-085, MBZIRC-8434000194, KU-BIT-Joint-Lab-8434000534 and KU-Stanford :8474000605.

\section*{Decleration}
During the preparation of this work the author(s) used ChatGPT-4 in order to improve language and readability. After using this tool/service, the author(s) reviewed and edited the content as needed and take(s) full responsibility for the content of the publication.

\printcredits

\bibliographystyle{model1-num-names}

\bibliography{cas-refs}

\begin{thebibliography}{63}
\expandafter\ifx\csname natexlab\endcsname\relax\def\natexlab#1{#1}\fi
\providecommand{\url}[1]{\texttt{#1}}
\providecommand{\href}[2]{#2}
\providecommand{\path}[1]{#1}
\providecommand{\DOIprefix}{doi:}
\providecommand{\ArXivprefix}{arXiv:}
\providecommand{\URLprefix}{URL: }
\providecommand{\Pubmedprefix}{pmid:}
\providecommand{\doi}[1]{\href{http://dx.doi.org/#1}{\path{#1}}}
\providecommand{\Pubmed}[1]{\href{pmid:#1}{\path{#1}}}
\providecommand{\bibinfo}[2]{#2}
\ifx\xfnm\relax \def\xfnm[#1]{\unskip,\space#1}\fi
\bibitem[{Subasinghe et~al.(2009)Subasinghe, Soto, and Jia}]{subasinghe2009global}
\bibinfo{author}{R.~Subasinghe}, \bibinfo{author}{D.~Soto}, \bibinfo{author}{J.~Jia},
\newblock \bibinfo{title}{Global aquaculture and its role in sustainable development},
\newblock \bibinfo{journal}{Reviews in aquaculture} \bibinfo{volume}{1} (\bibinfo{year}{2009}) \bibinfo{pages}{2--9}.
\bibitem[{Akram and Casavola(2021)}]{akram2021visual}
\bibinfo{author}{W.~Akram}, \bibinfo{author}{A.~Casavola},
\newblock \bibinfo{title}{A visual control scheme for auv underwater pipeline tracking},
\newblock in: \bibinfo{booktitle}{2021 IEEE International Conference on Autonomous Systems (ICAS)}, \bibinfo{organization}{IEEE}, \bibinfo{year}{2021}, pp. \bibinfo{pages}{1--5}.
\bibitem[{Paspalakis et~al.(2020)Paspalakis, Moirogiorgou, Papandroulakis, Giakos, and Zervakis}]{paspalakis2020automated}
\bibinfo{author}{S.~Paspalakis}, \bibinfo{author}{K.~Moirogiorgou}, \bibinfo{author}{N.~Papandroulakis}, \bibinfo{author}{G.~Giakos}, \bibinfo{author}{M.~Zervakis},
\newblock \bibinfo{title}{Automated fish cage net inspection using image processing techniques},
\newblock \bibinfo{journal}{IET Image Processing} \bibinfo{volume}{14} (\bibinfo{year}{2020}) \bibinfo{pages}{2028--2034}.
\bibitem[{Sohan et~al.(2024)Sohan, Sai~Ram, Reddy, and Venkata}]{sohan2024review}
\bibinfo{author}{M.~Sohan}, \bibinfo{author}{T.~Sai~Ram}, \bibinfo{author}{R.~Reddy}, \bibinfo{author}{C.~Venkata},
\newblock \bibinfo{title}{A review on yolov8 and its advancements},
\newblock in: \bibinfo{booktitle}{International Conference on Data Intelligence and Cognitive Informatics}, \bibinfo{organization}{Springer}, \bibinfo{year}{2024}, pp. \bibinfo{pages}{529--545}.
\bibitem[{Akram et~al.(2023)Akram, Ahmed, Seneviratne, and Hussain}]{akram2023autonomous}
\bibinfo{author}{W.~Akram}, \bibinfo{author}{M.~Ahmed}, \bibinfo{author}{L.~Seneviratne}, \bibinfo{author}{I.~Hussain},
\newblock \bibinfo{title}{Autonomous underwater robotic system for aquaculture applications},
\newblock \bibinfo{journal}{arXiv preprint arXiv:2308.14762}  (\bibinfo{year}{2023}).
\bibitem[{Salin and Arome~Ataguba(2018)}]{salin2018aquaculture}
\bibinfo{author}{K.~R. Salin}, \bibinfo{author}{G.~Arome~Ataguba},
\newblock \bibinfo{title}{Aquaculture and the environment: towards sustainability},
\newblock \bibinfo{journal}{Sustainable Aquaculture}  (\bibinfo{year}{2018}) \bibinfo{pages}{1--62}.
\bibitem[{Lee et~al.(2022)Lee, Jeong, Yu, and Ryu}]{lee2022autonomous}
\bibinfo{author}{H.~Lee}, \bibinfo{author}{D.~Jeong}, \bibinfo{author}{H.~Yu}, \bibinfo{author}{J.~Ryu},
\newblock \bibinfo{title}{Autonomous underwater vehicle control for fishnet inspection in turbid water environments},
\newblock \bibinfo{journal}{International Journal of Control, Automation and Systems} \bibinfo{volume}{20} (\bibinfo{year}{2022}) \bibinfo{pages}{3383--3392}.
\bibitem[{Achiam et~al.(2023)Achiam, Adler, Agarwal, Ahmad, Akkaya, Aleman, Almeida, Altenschmidt, Altman, Anadkat et~al.}]{achiam2023gpt}
\bibinfo{author}{J.~Achiam}, \bibinfo{author}{S.~Adler}, \bibinfo{author}{S.~Agarwal}, \bibinfo{author}{L.~Ahmad}, \bibinfo{author}{I.~Akkaya}, \bibinfo{author}{F.~L. Aleman}, \bibinfo{author}{D.~Almeida}, \bibinfo{author}{J.~Altenschmidt}, \bibinfo{author}{S.~Altman}, \bibinfo{author}{S.~Anadkat}, et~al.,
\newblock \bibinfo{title}{Gpt-4 technical report},
\newblock \bibinfo{journal}{arXiv preprint arXiv:2303.08774}  (\bibinfo{year}{2023}).
\bibitem[{Wang et~al.(2024)Wang, Ma, Feng, Zhang, Yang, Zhang, Chen, Tang, Chen, Lin et~al.}]{wang2024survey}
\bibinfo{author}{L.~Wang}, \bibinfo{author}{C.~Ma}, \bibinfo{author}{X.~Feng}, \bibinfo{author}{Z.~Zhang}, \bibinfo{author}{H.~Yang}, \bibinfo{author}{J.~Zhang}, \bibinfo{author}{Z.~Chen}, \bibinfo{author}{J.~Tang}, \bibinfo{author}{X.~Chen}, \bibinfo{author}{Y.~Lin}, et~al.,
\newblock \bibinfo{title}{A survey on large language model based autonomous agents},
\newblock \bibinfo{journal}{Frontiers of Computer Science} \bibinfo{volume}{18} (\bibinfo{year}{2024}) \bibinfo{pages}{186345}.
\bibitem[{Yang et~al.(2023)Yang, Hou, Wang, and Zhang}]{yang2023oceanchat}
\bibinfo{author}{R.~Yang}, \bibinfo{author}{M.~Hou}, \bibinfo{author}{J.~Wang}, \bibinfo{author}{F.~Zhang},
\newblock \bibinfo{title}{Oceanchat: Piloting autonomous underwater vehicles in natural language},
\newblock \bibinfo{journal}{arXiv preprint arXiv:2309.16052}  (\bibinfo{year}{2023}).
\bibitem[{Yang et~al.(2024)Yang, Zhang, and Hou}]{yang2024oceanplan}
\bibinfo{author}{R.~Yang}, \bibinfo{author}{F.~Zhang}, \bibinfo{author}{M.~Hou},
\newblock \bibinfo{title}{Oceanplan: Hierarchical planning and replanning for natural language auv piloting in large-scale unexplored ocean environments},
\newblock \bibinfo{journal}{arXiv preprint arXiv:2403.15369}  (\bibinfo{year}{2024}).
\bibitem[{Chen et~al.(2024)Chen, Blow, Abdullah, and Islam}]{chen2024word2wave}
\bibinfo{author}{R.~Chen}, \bibinfo{author}{D.~Blow}, \bibinfo{author}{A.~Abdullah}, \bibinfo{author}{M.~J. Islam},
\newblock \bibinfo{title}{Word2wave: Language driven mission programming for efficient subsea deployments of marine robots},
\newblock \bibinfo{journal}{arXiv preprint arXiv:2409.18405}  (\bibinfo{year}{2024}).
\bibitem[{Akram et~al.(2022)Akram, Casavola, and Mi{\v{s}}kovic}]{akram2022robust}
\bibinfo{author}{W.~Akram}, \bibinfo{author}{A.~Casavola}, \bibinfo{author}{N.~Mi{\v{s}}kovic},
\newblock \bibinfo{title}{Robust adaptive control allocation schemes for overactuated underwater vehicles under actuator faults},
\newblock \bibinfo{journal}{IFAC-PapersOnLine} \bibinfo{volume}{55} (\bibinfo{year}{2022}) \bibinfo{pages}{67--72}.
\bibitem[{Nguyen et~al.(2024)Nguyen, Caharija, Ohrem, Gravdahl, Loria, and Amundsen}]{nguyen2024robust}
\bibinfo{author}{K.~H. Nguyen}, \bibinfo{author}{W.~Caharija}, \bibinfo{author}{S.~J. Ohrem}, \bibinfo{author}{J.~T. Gravdahl}, \bibinfo{author}{A.~Loria}, \bibinfo{author}{H.~B. Amundsen},
\newblock \bibinfo{title}{Robust control of autonomous remotely operated vehicles at exposed aquaculture sites.}  (\bibinfo{year}{2024}).
\bibitem[{Ohrem et~al.(2022)Ohrem, Amundsen, Caharija, and Holden}]{ohrem2022robust}
\bibinfo{author}{S.~J. Ohrem}, \bibinfo{author}{H.~B. Amundsen}, \bibinfo{author}{W.~Caharija}, \bibinfo{author}{C.~Holden},
\newblock \bibinfo{title}{Robust adaptive backstepping dp control of rovs},
\newblock \bibinfo{journal}{Control Engineering Practice} \bibinfo{volume}{127} (\bibinfo{year}{2022}) \bibinfo{pages}{105282}.
\bibitem[{Ohrem et~al.(2023)Ohrem, Evjemo, Haugal{\o}kken, Amundsen, and Kelasidi}]{ohrem2023adaptive}
\bibinfo{author}{S.~J. Ohrem}, \bibinfo{author}{L.~D. Evjemo}, \bibinfo{author}{B.~O.~A. Haugal{\o}kken}, \bibinfo{author}{H.~B. Amundsen}, \bibinfo{author}{E.~Kelasidi},
\newblock \bibinfo{title}{Adaptive speed control of rovs with experimental results from an aquaculture net pen inspection operation},
\newblock in: \bibinfo{booktitle}{2023 31st Mediterranean Conference on Control and Automation (MED)}, \bibinfo{organization}{IEEE}, \bibinfo{year}{2023}, pp. \bibinfo{pages}{868--875}.
\bibitem[{Kelasidi et~al.(2022)Kelasidi, Su, Caharija, F{\o}re, Pedersen, and Frank}]{kelasidi2022autonomous}
\bibinfo{author}{E.~Kelasidi}, \bibinfo{author}{B.~Su}, \bibinfo{author}{W.~Caharija}, \bibinfo{author}{M.~F{\o}re}, \bibinfo{author}{M.~O. Pedersen}, \bibinfo{author}{K.~Frank},
\newblock \bibinfo{title}{Autonomous monitoring and inspection operations with uuvs in fish farms},
\newblock \bibinfo{journal}{IFAC-PapersOnLine} \bibinfo{volume}{55} (\bibinfo{year}{2022}) \bibinfo{pages}{401--408}.
\bibitem[{Ohrem et~al.(2024)Ohrem, Haugal{\o}kken, and Holden}]{ohrem2024application}
\bibinfo{author}{S.~J. Ohrem}, \bibinfo{author}{B.~O.~A. Haugal{\o}kken}, \bibinfo{author}{C.~Holden},
\newblock \bibinfo{title}{Application of modified model reference adaptive controller and observer (mraco) for speed control of an unmanned underwater vehicle},
\newblock \bibinfo{journal}{IFAC-PapersOnLine} \bibinfo{volume}{58} (\bibinfo{year}{2024}) \bibinfo{pages}{196--202}.
\bibitem[{Botta et~al.(2024)Botta, Ebner, Studer, Reijgwart, Siegwart, and Kelasidi}]{bott2024}
\bibinfo{author}{D.~Botta}, \bibinfo{author}{L.~Ebner}, \bibinfo{author}{A.~Studer}, \bibinfo{author}{V.~Reijgwart}, \bibinfo{author}{R.~Siegwart}, \bibinfo{author}{E.~Kelasidi}, \bibinfo{title}{Framework for robust localization of uuvs and mapping of net pens}, \bibinfo{year}{2024}. \URLprefix \url{https://arxiv.org/abs/2409.15475}. \href{http://arxiv.org/abs/2409.15475}{\tt arXiv:2409.15475}.
\bibitem[{Xia et~al.(2023)Xia, Ma, Li, Xu, and Qi}]{xia2023scale}
\bibinfo{author}{J.~Xia}, \bibinfo{author}{T.~Ma}, \bibinfo{author}{Y.~Li}, \bibinfo{author}{S.~Xu}, \bibinfo{author}{H.~Qi},
\newblock \bibinfo{title}{A scale-aware monocular odometry for fishnet inspection with both repeated and weak features},
\newblock \bibinfo{journal}{IEEE Transactions on Instrumentation and Measurement}  (\bibinfo{year}{2023}).
\bibitem[{Bjerkeng et~al.(2023)Bjerkeng, Gr{\o}tli, Kirkhus, Thielemann, Amundsen, Su, and Ohrem}]{bjerkeng2023absolute}
\bibinfo{author}{M.~Bjerkeng}, \bibinfo{author}{E.~I. Gr{\o}tli}, \bibinfo{author}{T.~Kirkhus}, \bibinfo{author}{J.~T. Thielemann}, \bibinfo{author}{H.~B. Amundsen}, \bibinfo{author}{B.~Su}, \bibinfo{author}{S.~Ohrem},
\newblock \bibinfo{title}{Absolute localization of an rov in a fish pen using laser triangulation},
\newblock in: \bibinfo{booktitle}{2023 31st Mediterranean Conference on Control and Automation (MED)}, \bibinfo{organization}{IEEE}, \bibinfo{year}{2023}, pp. \bibinfo{pages}{182--188}.
\bibitem[{Bjerkeng et~al.(2021)Bjerkeng, Kirkhus, Caharija, T.~Thielemann, B.~Amundsen, Johan~Ohrem, and Ingar~Gr{\o}tli}]{bjerkeng2021rov}
\bibinfo{author}{M.~Bjerkeng}, \bibinfo{author}{T.~Kirkhus}, \bibinfo{author}{W.~Caharija}, \bibinfo{author}{J.~T.~Thielemann}, \bibinfo{author}{H.~B.~Amundsen}, \bibinfo{author}{S.~Johan~Ohrem}, \bibinfo{author}{E.~Ingar~Gr{\o}tli},
\newblock \bibinfo{title}{Rov navigation in a fish cage with laser-camera triangulation},
\newblock \bibinfo{journal}{Journal of Marine Science and Engineering} \bibinfo{volume}{9} (\bibinfo{year}{2021}) \bibinfo{pages}{79}.
\bibitem[{Cardaillac et~al.(2023)Cardaillac, Amundsen, Kelasidi, and Ludvigsen}]{cardaillac2023application}
\bibinfo{author}{A.~Cardaillac}, \bibinfo{author}{H.~B. Amundsen}, \bibinfo{author}{E.~Kelasidi}, \bibinfo{author}{M.~Ludvigsen},
\newblock \bibinfo{title}{Application of maneuvering based control for autonomous inspection of aquaculture net pens},
\newblock in: \bibinfo{booktitle}{2023 8th Asia-Pacific Conference on Intelligent Robot Systems (ACIRS)}, \bibinfo{organization}{IEEE}, \bibinfo{year}{2023}, pp. \bibinfo{pages}{44--51}.
\bibitem[{Cardaillac et~al.(2024)Cardaillac, Skjetne, and Ludvigsen}]{cardaillac2024rov}
\bibinfo{author}{A.~Cardaillac}, \bibinfo{author}{R.~Skjetne}, \bibinfo{author}{M.~Ludvigsen},
\newblock \bibinfo{title}{Rov-based autonomous maneuvering for ship hull inspection with coverage monitoring},
\newblock \bibinfo{journal}{Journal of Intelligent \& Robotic Systems} \bibinfo{volume}{110} (\bibinfo{year}{2024}) \bibinfo{pages}{59}.
\bibitem[{Skaldeb{\o} et~al.(2024)Skaldeb{\o}, Schellewald, Evjemo, Amundsen, Xanthidis, and Kelasidi}]{skaldebo2024approaches}
\bibinfo{author}{M.~Skaldeb{\o}}, \bibinfo{author}{C.~Schellewald}, \bibinfo{author}{L.~D. Evjemo}, \bibinfo{author}{H.~B. Amundsen}, \bibinfo{author}{M.~Xanthidis}, \bibinfo{author}{E.~Kelasidi},
\newblock \bibinfo{title}{Approaches enabling underwater autonomy and sensing in sea-based aquaculture settings},
\newblock in: \bibinfo{booktitle}{2024 32nd Mediterranean Conference on Control and Automation (MED)}, \bibinfo{organization}{IEEE}, \bibinfo{year}{2024}, pp. \bibinfo{pages}{197--202}.
\bibitem[{Schellewald et~al.(2021)Schellewald, Stahl, and Kelasidi}]{schellewald2021vision}
\bibinfo{author}{C.~Schellewald}, \bibinfo{author}{A.~Stahl}, \bibinfo{author}{E.~Kelasidi},
\newblock \bibinfo{title}{Vision-based pose estimation for autonomous operations in aquacultural fish farms},
\newblock \bibinfo{journal}{IFAC-PapersOnLine} \bibinfo{volume}{54} (\bibinfo{year}{2021}) \bibinfo{pages}{438--443}.
\bibitem[{Wu et~al.(2022)Wu, Liu, Wei, An, Duan, Li, Li, Chen, and Wei}]{wu2022intelligent}
\bibinfo{author}{Y.~Wu}, \bibinfo{author}{J.~Liu}, \bibinfo{author}{Y.~Wei}, \bibinfo{author}{D.~An}, \bibinfo{author}{Y.~Duan}, \bibinfo{author}{W.~Li}, \bibinfo{author}{B.~Li}, \bibinfo{author}{Y.~Chen}, \bibinfo{author}{Q.~Wei},
\newblock \bibinfo{title}{Intelligent control method of underwater inspection robot in netcage},
\newblock \bibinfo{journal}{Aquaculture Research} \bibinfo{volume}{53} (\bibinfo{year}{2022}) \bibinfo{pages}{1928--1938}.
\bibitem[{Amundsen et~al.(2021)Amundsen, Caharija, and Pettersen}]{amundsen2021autonomous}
\bibinfo{author}{H.~B. Amundsen}, \bibinfo{author}{W.~Caharija}, \bibinfo{author}{K.~Y. Pettersen},
\newblock \bibinfo{title}{Autonomous rov inspections of aquaculture net pens using dvl},
\newblock \bibinfo{journal}{IEEE Journal of Oceanic Engineering} \bibinfo{volume}{47} (\bibinfo{year}{2021}) \bibinfo{pages}{1--19}.
\bibitem[{Rosa et~al.(2024)Rosa, Cabecinhas, and Ferreira}]{rosa2024forward}
\bibinfo{author}{D.~Rosa}, \bibinfo{author}{D.~Cabecinhas}, \bibinfo{author}{F.~Ferreira},
\newblock \bibinfo{title}{Forward-looking sonar based autonomous aquaculture inspection},
\newblock in: \bibinfo{booktitle}{OCEANS 2024-Singapore}, \bibinfo{organization}{IEEE}, \bibinfo{year}{2024}, pp. \bibinfo{pages}{1--8}.
\bibitem[{Tun et~al.(2023)Tun, Huang, and Preece}]{tun2023development}
\bibinfo{author}{T.~T. Tun}, \bibinfo{author}{L.~Huang}, \bibinfo{author}{M.~A. Preece},
\newblock \bibinfo{title}{Development and high-fidelity simulation of trajectory tracking control schemes of a uuv for fish net-pen visual inspection in offshore aquaculture},
\newblock \bibinfo{journal}{IEEE Access} \bibinfo{volume}{11} (\bibinfo{year}{2023}) \bibinfo{pages}{135764--135787}.
\bibitem[{Akram et~al.(2022)Akram, Casavola, Kapetanovic, and Miskovic}]{akram2022visual}
\bibinfo{author}{W.~Akram}, \bibinfo{author}{A.~Casavola}, \bibinfo{author}{N.~Kapetanovic}, \bibinfo{author}{N.~Miskovic},
\newblock \bibinfo{title}{A visual servoing scheme for autonomous aquaculture net pens inspection using rov},
\newblock \bibinfo{journal}{Sensors} \bibinfo{volume}{22} (\bibinfo{year}{2022}) \bibinfo{pages}{3525}.
\bibitem[{Haugal{\o}kken et~al.(2024)Haugal{\o}kken, Nissen, Skaldeb{\o}, Ohrem, and Kelasidi}]{haugalokken2024low}
\bibinfo{author}{B.~O. Haugal{\o}kken}, \bibinfo{author}{O.~Nissen}, \bibinfo{author}{M.~B. Skaldeb{\o}}, \bibinfo{author}{S.~J. Ohrem}, \bibinfo{author}{E.~Kelasidi},
\newblock \bibinfo{title}{Low-cost sensor technologies for underwater vehicle navigation in aquaculture net pens},
\newblock \bibinfo{journal}{IFAC-PapersOnLine} \bibinfo{volume}{58} (\bibinfo{year}{2024}) \bibinfo{pages}{87--94}.
\bibitem[{Rahim et~al.(2022)Rahim, Hamid, Said, Jamaludin, and Rozalan}]{rahim2022design}
\bibinfo{author}{M.~N.~A. Rahim}, \bibinfo{author}{D.~T.~A. Hamid}, \bibinfo{author}{M.~F.~M. Said}, \bibinfo{author}{M.~N. Jamaludin}, \bibinfo{author}{H.~H. Rozalan},
\newblock \bibinfo{title}{Design and development of remotely operated vehicle to assess the water quality in aquaculture area},
\newblock \bibinfo{journal}{Asian People Journal (APJ)} \bibinfo{volume}{5} (\bibinfo{year}{2022}) \bibinfo{pages}{50--60}.
\bibitem[{Tarwadi et~al.(2020)Tarwadi, Shiraki, Ganoni, Wei, Ahn, and MacDonald}]{tarwadi2020design}
\bibinfo{author}{P.~Tarwadi}, \bibinfo{author}{Y.~Shiraki}, \bibinfo{author}{O.~Ganoni}, \bibinfo{author}{S.~Wei}, \bibinfo{author}{H.~S. Ahn}, \bibinfo{author}{B.~MacDonald},
\newblock \bibinfo{title}{Design and development of a robotic vehicle for shallow-water marine inspections},
\newblock \bibinfo{journal}{arXiv preprint arXiv:2007.04563}  (\bibinfo{year}{2020}).
\bibitem[{Vasileiou et~al.(2021)Vasileiou, Manos, and Kavallieratou}]{vasileiou2021low}
\bibinfo{author}{M.~Vasileiou}, \bibinfo{author}{N.~Manos}, \bibinfo{author}{E.~Kavallieratou},
\newblock \bibinfo{title}{A low-cost 3d printed mini underwater vehicle: Design and fabrication},
\newblock in: \bibinfo{booktitle}{2021 20th International Conference on Advanced Robotics (ICAR)}, \bibinfo{organization}{IEEE}, \bibinfo{year}{2021}, pp. \bibinfo{pages}{390--395}.
\bibitem[{Xu et~al.(2023)Xu, Zhou, Yuan, Guo, Huang, and Zhang}]{xu2023vision}
\bibinfo{author}{G.~Xu}, \bibinfo{author}{D.~Zhou}, \bibinfo{author}{L.~Yuan}, \bibinfo{author}{W.~Guo}, \bibinfo{author}{Z.~Huang}, \bibinfo{author}{Y.~Zhang},
\newblock \bibinfo{title}{Vision-based underwater target real-time detection for autonomous underwater vehicle subsea exploration},
\newblock \bibinfo{journal}{Frontiers in Marine Science} \bibinfo{volume}{10} (\bibinfo{year}{2023}) \bibinfo{pages}{1112310}.
\bibitem[{Akram et~al.(2025)Akram, Baidar~Bakht, Ud~Din, Seneviratne, and Hussain}]{aquadata}
\bibinfo{author}{W.~Akram}, \bibinfo{author}{A.~Baidar~Bakht}, \bibinfo{author}{M.~Ud~Din}, \bibinfo{author}{L.~Seneviratne}, \bibinfo{author}{I.~Hussain},
\newblock \bibinfo{title}{Enhancing aquaculture net pen inspection: A benchmark study on detection and semantic segmentation},
\newblock \bibinfo{journal}{IEEE Access} \bibinfo{volume}{13} (\bibinfo{year}{2025}) \bibinfo{pages}{3453--3474}.
\bibitem[{Akram et~al.(2024)Akram, Hassan, Toubar, Ahmed, Mi{\v{s}}kovic, Seneviratne, and Hussain}]{akram2024aquaculture}
\bibinfo{author}{W.~Akram}, \bibinfo{author}{T.~Hassan}, \bibinfo{author}{H.~Toubar}, \bibinfo{author}{M.~Ahmed}, \bibinfo{author}{N.~Mi{\v{s}}kovic}, \bibinfo{author}{L.~Seneviratne}, \bibinfo{author}{I.~Hussain},
\newblock \bibinfo{title}{Aquaculture defects recognition via multi-scale semantic segmentation},
\newblock \bibinfo{journal}{Expert systems with applications} \bibinfo{volume}{237} (\bibinfo{year}{2024}) \bibinfo{pages}{121197}.
\bibitem[{Akram et~al.(2023)Akram, Ahmed, Seneviratne, and Hussain}]{akram2023evaluating}
\bibinfo{author}{W.~Akram}, \bibinfo{author}{M.~Ahmed}, \bibinfo{author}{L.~Seneviratne}, \bibinfo{author}{I.~Hussain},
\newblock \bibinfo{title}{Evaluating deep learning assisted automated aquaculture net pens inspection using rov},
\newblock \bibinfo{journal}{arXiv preprint arXiv:2308.13826}  (\bibinfo{year}{2023}).
\bibitem[{Madshaven et~al.(2022)Madshaven, Schellewald, and Stahl}]{madshaven2022hole}
\bibinfo{author}{A.~Madshaven}, \bibinfo{author}{C.~Schellewald}, \bibinfo{author}{A.~Stahl},
\newblock \bibinfo{title}{Hole detection in aquaculture net cages from video footage},
\newblock in: \bibinfo{booktitle}{Fourteenth International Conference on Machine Vision (ICMV 2021)}, volume \bibinfo{volume}{12084}, \bibinfo{organization}{SPIE}, \bibinfo{year}{2022}, pp. \bibinfo{pages}{258--267}.
\bibitem[{Schellewald and Stahl(2022)}]{schellewald2022irregularity}
\bibinfo{author}{C.~Schellewald}, \bibinfo{author}{A.~Stahl},
\newblock \bibinfo{title}{Irregularity detection in net pens exploiting computer vision},
\newblock \bibinfo{journal}{IFAC-PapersOnLine} \bibinfo{volume}{55} (\bibinfo{year}{2022}) \bibinfo{pages}{415--420}.
\bibitem[{Ronneberger et~al.(2015)Ronneberger, Fischer, and Brox}]{unet}
\bibinfo{author}{O.~Ronneberger}, \bibinfo{author}{P.~Fischer}, \bibinfo{author}{T.~Brox},
\newblock \bibinfo{title}{U-net: Convolutional networks for biomedical image segmentation},
\newblock in: \bibinfo{booktitle}{Medical image computing and computer-assisted intervention--MICCAI 2015: 18th international conference, Munich, Germany, October 5-9, 2015, proceedings, part III 18}, \bibinfo{organization}{Springer}, \bibinfo{year}{2015}, pp. \bibinfo{pages}{234--241}.
\bibitem[{Zacheilas et~al.(2021)Zacheilas, Moirogiorgou, Papandroulakis, Sotiriades, Zervakis, and Dollas}]{zacheilas2021fpga}
\bibinfo{author}{T.~Zacheilas}, \bibinfo{author}{K.~Moirogiorgou}, \bibinfo{author}{N.~Papandroulakis}, \bibinfo{author}{E.~Sotiriades}, \bibinfo{author}{M.~Zervakis}, \bibinfo{author}{A.~Dollas},
\newblock \bibinfo{title}{An fpga-based system for video processing to detect holes in aquaculture nets},
\newblock in: \bibinfo{booktitle}{2021 IEEE 21st International Conference on Bioinformatics and Bioengineering (BIBE)}, \bibinfo{organization}{IEEE}, \bibinfo{year}{2021}, pp. \bibinfo{pages}{1--6}.
\bibitem[{Jocher et~al.(2020)Jocher, Chaurasia, Stoken, and et~al.}]{yolov5}
\bibinfo{author}{G.~Jocher}, \bibinfo{author}{A.~Chaurasia}, \bibinfo{author}{A.~Stoken}, \bibinfo{author}{et~al.}, \bibinfo{title}{Yolov5 - you only look once}, \bibinfo{howpublished}{\url{https://github.com/ultralytics/yolov5}}, \bibinfo{year}{2020}. \bibinfo{note}{Accessed: 2024-09-23}.
\bibitem[{Paraskevas and Kavallieratou(2023)}]{paraskevas2023detecting}
\bibinfo{author}{K.~Paraskevas}, \bibinfo{author}{E.~Kavallieratou},
\newblock \bibinfo{title}{Detecting holes in fish farming nets: A two--method approach},
\newblock in: \bibinfo{booktitle}{2023 International Conference on Control, Automation and Diagnosis (ICCAD)}, \bibinfo{organization}{IEEE}, \bibinfo{year}{2023}, pp. \bibinfo{pages}{1--7}.
\bibitem[{Paraskevas and Kavallieratou(2022)}]{paraskevas2022detecting}
\bibinfo{author}{K.~Paraskevas}, \bibinfo{author}{E.~Kavallieratou},
\newblock \bibinfo{title}{Detecting holes in fishery nets using an rov},
\newblock in: \bibinfo{booktitle}{2022 International Conference on Electrical, Computer, Communications and Mechatronics Engineering (ICECCME)}, \bibinfo{organization}{IEEE}, \bibinfo{year}{2022}, pp. \bibinfo{pages}{1--5}.
\bibitem[{Kang et~al.(2023)Kang, Keruzel, Baek, and Lee}]{kang2023detection}
\bibinfo{author}{J.-H. Kang}, \bibinfo{author}{T.~Keruzel}, \bibinfo{author}{U.-J. Baek}, \bibinfo{author}{K.-C. Lee},
\newblock \bibinfo{title}{Detection of fish cage net damage using image processing with mesh-hole grouping},
\newblock in: \bibinfo{booktitle}{2023 IEEE Region 10 Symposium (TENSYMP)}, \bibinfo{organization}{IEEE}, \bibinfo{year}{2023}, pp. \bibinfo{pages}{1--3}.
\bibitem[{Qiu et~al.(2020)Qiu, Pakrashi, and Ghosh}]{qiu2020fishing}
\bibinfo{author}{W.~Qiu}, \bibinfo{author}{V.~Pakrashi}, \bibinfo{author}{B.~Ghosh},
\newblock \bibinfo{title}{Fishing net health state estimation using underwater imaging},
\newblock \bibinfo{journal}{Journal of Marine Science and Engineering} \bibinfo{volume}{8} (\bibinfo{year}{2020}) \bibinfo{pages}{707}.
\bibitem[{Zhang et~al.(2022)Zhang, Gui, Qu, and Feng}]{zhang2022netting}
\bibinfo{author}{Z.~Zhang}, \bibinfo{author}{F.~Gui}, \bibinfo{author}{X.~Qu}, \bibinfo{author}{D.~Feng},
\newblock \bibinfo{title}{Netting damage detection for marine aquaculture facilities based on improved mask r-cnn},
\newblock \bibinfo{journal}{Journal of Marine Science and Engineering} \bibinfo{volume}{10} (\bibinfo{year}{2022}) \bibinfo{pages}{996}.
\bibitem[{L{\'o}pez-Barajas et~al.(2023)L{\'o}pez-Barajas, Sanz, Mar{\'\i}n-Prades, G{\'o}mez-Espinosa, Gonz{\'a}lez-Garc{\'\i}a, and Echag{\"u}e}]{lopez2023inspection}
\bibinfo{author}{S.~L{\'o}pez-Barajas}, \bibinfo{author}{P.~J. Sanz}, \bibinfo{author}{R.~Mar{\'\i}n-Prades}, \bibinfo{author}{A.~G{\'o}mez-Espinosa}, \bibinfo{author}{J.~Gonz{\'a}lez-Garc{\'\i}a}, \bibinfo{author}{J.~Echag{\"u}e},
\newblock \bibinfo{title}{Inspection operations and hole detection in fish net cages through a hybrid underwater intervention system using deep learning techniques},
\newblock \bibinfo{journal}{Journal of Marine Science and Engineering} \bibinfo{volume}{12} (\bibinfo{year}{2023}) \bibinfo{pages}{80}.
\bibitem[{Betancourt et~al.(2020)Betancourt, Coral, and Colorado}]{betancourt2020integrated}
\bibinfo{author}{J.~Betancourt}, \bibinfo{author}{W.~Coral}, \bibinfo{author}{J.~Colorado},
\newblock \bibinfo{title}{An integrated rov solution for underwater net-cage inspection in fish farms using computer vision},
\newblock \bibinfo{journal}{SN Applied Sciences} \bibinfo{volume}{2} (\bibinfo{year}{2020}) \bibinfo{pages}{1946}.
\bibitem[{Bi et~al.(2023)Bi, Zhang, Xue, Ou, Ji, Zheng, and Chen}]{bi2023oceangpt}
\bibinfo{author}{Z.~Bi}, \bibinfo{author}{N.~Zhang}, \bibinfo{author}{Y.~Xue}, \bibinfo{author}{Y.~Ou}, \bibinfo{author}{D.~Ji}, \bibinfo{author}{G.~Zheng}, \bibinfo{author}{H.~Chen},
\newblock \bibinfo{title}{Oceangpt: A large language model for ocean science tasks},
\newblock \bibinfo{journal}{arXiv preprint arXiv:2310.02031}  (\bibinfo{year}{2023}).
\bibitem[{Samuel et~al.(2024)Samuel, Sermet, Cwiertny, and Demir}]{samuel2024integrating}
\bibinfo{author}{D.~J. Samuel}, \bibinfo{author}{Y.~Sermet}, \bibinfo{author}{D.~Cwiertny}, \bibinfo{author}{I.~Demir},
\newblock \bibinfo{title}{Integrating vision-based ai and large language models for real-time water pollution surveillance},
\newblock \bibinfo{journal}{Water Environment Research} \bibinfo{volume}{96} (\bibinfo{year}{2024}) \bibinfo{pages}{e11092}.
\bibitem[{Khanal et~al.(2024)Khanal, Yu, Chiu, Chaudhary, Zhang, Katija, and Forbes}]{khanal2024fathomgpt}
\bibinfo{author}{N.~Khanal}, \bibinfo{author}{C.~M. Yu}, \bibinfo{author}{J.-C. Chiu}, \bibinfo{author}{A.~Chaudhary}, \bibinfo{author}{Z.~Zhang}, \bibinfo{author}{K.~Katija}, \bibinfo{author}{A.~G. Forbes},
\newblock \bibinfo{title}{Fathomgpt: A natural language interface for interactively exploring ocean science data},
\newblock in: \bibinfo{booktitle}{Proceedings of the 37th Annual ACM Symposium on User Interface Software and Technology}, \bibinfo{year}{2024}, pp. \bibinfo{pages}{1--15}.
\bibitem[{Lian and Li(2024)}]{lian2024evaluation}
\bibinfo{author}{S.~Lian}, \bibinfo{author}{H.~Li},
\newblock \bibinfo{title}{Evaluation of segment anything model 2: The role of sam2 in the underwater environment},
\newblock \bibinfo{journal}{arXiv preprint arXiv:2408.02924}  (\bibinfo{year}{2024}).
\bibitem[{Cahyadi et~al.(2023)Cahyadi, Asfihani, Mardiyanto, and Erfianti}]{cahyadi2023performance}
\bibinfo{author}{M.~N. Cahyadi}, \bibinfo{author}{T.~Asfihani}, \bibinfo{author}{R.~Mardiyanto}, \bibinfo{author}{R.~Erfianti},
\newblock \bibinfo{title}{Performance of gps and imu sensor fusion using unscented kalman filter for precise i-boat navigation in infinite wide waters},
\newblock \bibinfo{journal}{Geodesy and Geodynamics} \bibinfo{volume}{14} (\bibinfo{year}{2023}) \bibinfo{pages}{265--274}.
\bibitem[{Ang et~al.(2005)Ang, Chong, and Li}]{ang2005pid}
\bibinfo{author}{K.~H. Ang}, \bibinfo{author}{G.~Chong}, \bibinfo{author}{Y.~Li},
\newblock \bibinfo{title}{Pid control system analysis, design, and technology},
\newblock \bibinfo{journal}{IEEE transactions on control systems technology} \bibinfo{volume}{13} (\bibinfo{year}{2005}) \bibinfo{pages}{559--576}.
\bibitem[{Team(2018)}]{uuv_simulator}
\bibinfo{author}{U.~S.~D. Team}, \bibinfo{title}{Uuv simulator: Gazebo/ros packages for underwater vehicle simulation}, \bibinfo{howpublished}{\\url{https://uuvsimulator.github.io}}, \bibinfo{year}{2018}. \bibinfo{note}{Accessed: 2025-01-16}.
\bibitem[{Robotics and Contributors(2018)}]{bluerov2_simulator}
\bibinfo{author}{B.~Robotics}, \bibinfo{author}{Contributors}, \bibinfo{title}{Bluerov2 simulator: Simulation package for bluerov2 in ros/gazebo}, \bibinfo{howpublished}{\\url{https://github.com/HKPolyU-UAV/bluerov2}}, \bibinfo{year}{2018}. \bibinfo{note}{Accessed: 2025-01-16}.
\bibitem[{Robotics(2025)}]{BlueROV2}
\bibinfo{author}{B.~Robotics}, \bibinfo{title}{Bluerov2 - affordable and capable underwater rov}, \bibinfo{year}{2025}. \URLprefix \url{https://bluerobotics.com/store/rov/bluerov2/}, \bibinfo{note}{accessed: 2025-01-16}.
\bibitem[{von Benzon et~al.(2022)von Benzon, S{\o}rensen, Uth, Jouffroy, Liniger, and Pedersen}]{von2022open}
\bibinfo{author}{M.~von Benzon}, \bibinfo{author}{F.~F. S{\o}rensen}, \bibinfo{author}{E.~Uth}, \bibinfo{author}{J.~Jouffroy}, \bibinfo{author}{J.~Liniger}, \bibinfo{author}{S.~Pedersen},
\newblock \bibinfo{title}{An open-source benchmark simulator: Control of a bluerov2 underwater robot},
\newblock \bibinfo{journal}{Journal of Marine Science and Engineering} \bibinfo{volume}{10} (\bibinfo{year}{2022}) \bibinfo{pages}{1898}.
\bibitem[{Robotics(2025{\natexlab{a}})}]{blueye_x3}
\bibinfo{author}{B.~Robotics}, \bibinfo{title}{Blueye x3 rov}, \bibinfo{howpublished}{\url{https://www.blueyerobotics.com/products/x3}}, \bibinfo{year}{2025}{\natexlab{a}}. \bibinfo{note}{Accessed: 2025-01-23}.
\bibitem[{Robotics(2025{\natexlab{b}})}]{blueye_sdk}
\bibinfo{author}{B.~Robotics}, \bibinfo{title}{Blueye robotics sdk}, \bibinfo{howpublished}{\url{https://github.com/BluEye-Robotics/blueye.sdk}}, \bibinfo{year}{2025}{\natexlab{b}}. \bibinfo{note}{Accessed: 2025-01-23}.

\end{thebibliography}


\end{document}